\documentclass[10pt]{article}
\usepackage[letterpaper,top=0.55in,bottom=0.65in,left=0.58in,right=0.58in]{geometry}
\usepackage[T1]{fontenc}
\usepackage[utf8]{inputenc}
\usepackage[scaled=0.95]{helvet}

\usepackage{microtype}
\usepackage{graphicx}
\usepackage{amsmath,amssymb}
\usepackage{tabularx,longtable,array,booktabs}
\usepackage{enumitem}
\usepackage{float}
\usepackage{dblfloatfix}
\usepackage{caption}
\usepackage{xcolor}
\usepackage{placeins}
\usepackage{titlesec}
\definecolor{JMIRBlue}{HTML}{005B96}
\usepackage[colorlinks=true,linkcolor=JMIRBlue,urlcolor=JMIRBlue,citecolor=JMIRBlue,bookmarksopen=true,bookmarksnumbered=false]{hyperref}
\usepackage[all]{hypcap}
\usepackage{bookmark}
\titleformat{\section}{\fontsize{15}{17}\selectfont\bfseries}{ }{0pt}{}
\titleformat{\subsection}{\fontsize{11.2}{13}\selectfont\bfseries}{ }{0pt}{}
\titlespacing*{\section}{0pt}{9pt}{4pt}
\titlespacing*{\subsection}{0pt}{7pt}{3pt}
\setlist{nosep,leftmargin=*}
\makeatletter
\renewcommand{\maketitle}{%
  \begin{flushleft}
  {\fontsize{19}{21}\selectfont\bfseries \@title\par}
  \vspace{7pt}
  {\fontsize{10}{12}\selectfont \@author\par}
  \vspace{6pt}
  \end{flushleft}
}
\makeatother
\title{SkinAgent AI: A Safety-Grounded Multimodal Agentic Framework for Non-Diagnostic Skincare Support}

\author{%
Muhammad Muhtasim Shahriar\textsuperscript{1},
Abdullah Mohammad Sayem\textsuperscript{1},
Liew Tze Hui\textsuperscript{2,*},
M.\,F. Mridha\textsuperscript{3},
Md. Mahiuddin\textsuperscript{1}
\\[3pt]
{\small
\textsuperscript{1}Department of Computer Science, International Islamic University Chittagong, Chittagong, Bangladesh
\\
\textsuperscript{2}Centre for Intelligent Cloud Computing, Centre of Excellence for Advanced Cloud, Faculty of Information Science and Technology, Multimedia University, Melaka, Malaysia
\\
\textsuperscript{3}Department of Computer Science, American International University-Bangladesh, Dhaka, Bangladesh
}
\\[3pt]
{\small
\textsuperscript{*}Corresponding author: Liew Tze Hui, PhD;
Multimedia University, Jalan Ayer Keroh Lama, 75450 Bukit Beruang, Melaka, Malaysia;
\href{mailto:thliew@mmu.edu.my}{thliew@mmu.edu.my}
}
}

\date{}
\begin{document}
\twocolumn[
\begin{@twocolumnfalse}
\maketitle
\begin{center}\textbf{Original Paper}\end{center}
\section*{Abstract}

\textbf{Background:} Consumer-facing skincare AI must coordinate visual evidence, product information, tool use, and user-facing actions. These components require explicit evidence boundaries and safety controls that cannot be provided by isolated image classification or unconstrained language generation alone.

\textbf{Objective:} This study evaluates SkinAgent AI, a non-diagnostic multimodal framework that combines visual concern routing with grounded and auditable orchestration supported by a large language model.

\textbf{Methods:} The framework routes images to acne, pore, and wrinkle analysis modules and includes supporting skin-type estimation and count-informed ordinal acne-severity assessment. Its agent layer uses typed tools, database-grounded recommendation and action functions, deterministic safety, privacy, and evidence checks, approval before action execution, and structured trace and replay mechanisms. Visual-model performance and system-level agent behavior were evaluated separately.

\textbf{Results:} The final internal condition experiment produced an accuracy of $99.84\%\pm0.07\%$ across three seeds. The skin-type module achieved 88.85\% accuracy, while count-informed acne-severity assessment achieved 84.59\% accuracy with a quadratic weighted kappa of 0.9076. On a locked but non-independent system benchmark of 240 cases, intent accuracy was 80.00\%, exact tool-set match was 62.92\%, and strict task completion was 47.08\%. These results show a substantial difference between recognizing a request and executing it reliably. No violations or successful cross-user leakage events were observed in the finite safety and privacy test suites. Tool-selection errors, incomplete grounding of product attributes, and unreliable failure fallback nevertheless remained. A controlled comparison of system configurations also showed endpoint-specific trade-offs; the most constrained configuration did not perform best on every measure.

\textbf{Conclusions:} The findings support the feasibility of combining visual AI with bounded, database-grounded, and traceable agent orchestration for non-diagnostic skincare support. They do not establish clinical readiness, external generalization, formal privacy guarantees, or universal safety. Independent validation using multiple data sources, expert assessment, robustness and fairness testing, and prospective evaluation in real-world settings remain necessary.

\textbf{Keywords: Agentic AI; Multimodal AI; Skincare AI; Tool-using large language models; Visual concern routing; Database grounding; Safety-grounded AI; Auditable AI}

\vspace{4pt}
\end{@twocolumnfalse}
]

\section{Introduction}

Artificial intelligence is increasingly used in dermatology and cosmetic skin analysis because visible attributes, lesion morphology, texture, and spatial distribution can be captured from images and processed at scale. Recent reviews cover applications in skin assessment, skin-condition analysis, product support, and treatment-outcome prediction, while dermatological research continues to rely heavily on image-analysis pipelines [\hyperlink{ref:1}{1}], [\hyperlink{ref:2}{2}]. Acne is a representative image-based problem. Existing studies address lesion detection, counting, segmentation, and severity grading, reflecting the importance of both local lesion evidence and global facial appearance [\hyperlink{ref:3}{3}], [\hyperlink{ref:4}{4}].

Smartphone applications and web-based skin services have extended these capabilities beyond specialist settings. Although such systems can make support more accessible, they also raise questions about intended use, data quality, privacy, and user-facing safeguards [\hyperlink{ref:5}{5}]. A consumer-facing system must therefore interpret visible concerns, preserve uncertainty, and provide structured support without presenting its output as a diagnosis or treatment decision. Its quality depends on more than classification accuracy. The system must also define which evidence it may use, which actions it may perform, and how it communicates uncertainty and risk.

Vision models and conversational models address different parts of this problem. Dermatology studies have observed that unimodal systems are limited by their reliance on a single source of information, whereas multimodal approaches can combine images with textual or contextual inputs [\hyperlink{ref:6}{6}]. Acne-assessment frameworks have combined global and local image features or distilled lesion-level evidence into severity models, but these systems remain focused on visual prediction rather than a complete user-support workflow [\hyperlink{ref:7}{7}], [\hyperlink{ref:8}{8}].

Research on agents based on large language models (LLMs) commonly describes autonomous systems in terms of planning and action modules. Executable planning can nevertheless produce unnecessary or conflicting action sequences when the system lacks sufficient knowledge about the available actions [\hyperlink{ref:9}{9}], [\hyperlink{ref:10}{10}]. Tool-augmented language models can call external application programming interfaces, and large-scale e-commerce search systems show the value of retrieving product information from structured catalogs [\hyperlink{ref:11}{11}], [\hyperlink{ref:12}{12}]. In facial applications, this integration also creates privacy risks. Face-processing systems may expose identity-related information or accumulate sensitive visual data unless storage, access, and output boundaries are explicitly controlled [\hyperlink{ref:13}{13}]. A practical skincare assistant therefore requires mechanisms that ground generated content in trusted evidence and prevent sensitive or unsupported information from reaching the user.

Work on visual perception, multimodal context, tool use, and process evaluation has largely progressed along separate paths. Structured agent frameworks show that role-specific components and explicit communication procedures can coordinate multistep tasks in bounded domains. Recent evaluations of agentic systems have also compared controlled system configurations using process-level measures [\hyperlink{ref:14}{14}], [\hyperlink{ref:15}{15}]. Dermatological deployment introduces additional concerns. Underrepresentation of skin of color, variation in image quality, and limited standardization may lead to uneven performance, while generalization from controlled studies to routine use remains uncertain [\hyperlink{ref:16}{16}], [\hyperlink{ref:17}{17}]. Acne research has advanced from joint grading and counting to lesion segmentation and scoring, but most studies still address a bounded visual-analysis task rather than the controlled use of visual evidence, recommendations, and user actions within one system [\hyperlink{ref:4}{4}], [\hyperlink{ref:18}{18}].

The remaining gap is therefore at the system level. Existing work does not adequately show how one auditable, non-diagnostic skincare framework can combine visual concern routing, bounded agent planning, typed tool use, database-grounded product evidence, deterministic safety and privacy checks, approval-controlled actions, and replayable execution traces. These components must also be evaluated at both the model and process levels rather than solely through the apparent quality of the final response [\hyperlink{ref:5}{5}], [\hyperlink{ref:15}{15}], [\hyperlink{ref:17}{17}].

To address this gap, this paper presents SkinAgent AI, a safety-grounded multimodal agentic framework for non-diagnostic skincare support. The system uses visual analysis as structured evidence within an orchestrated workflow rather than treating image classification as the final output. It combines input-quality assessment and facial-image analysis with concern-specific routing for acne, pores, and wrinkles. Supporting components estimate a skin-type category and provide count-informed ordinal acne-severity support through a dedicated Count-Informed Ordinal Severity (CIOS) module when acne analysis is appropriate.

An LLM-based orchestrator interprets the user's request and coordinates approved tools. Catalog retrieval and validation restrict product statements to evidence stored in the database. Deterministic checks for safety, privacy, evidence completeness, and action approval operate around the generative component. A response finalizer separates user-facing content from restricted expert and audit metadata. Structured execution traces record routing decisions, tool calls, verification results, and the finalization path, allowing the workflow to be inspected and replayed. This separation gives the conversational component flexibility while explicit rules and grounded data constrain evidence-sensitive claims and higher-risk actions.

The main contributions of this work are as follows:

\begin{enumerate}
    \item \textbf{Bounded multimodal orchestration:} We introduce a workflow organized around planning, execution, grounding, verification, finalization, and tracing. The LLM operates within explicit tool, evidence, and verification boundaries.

    \item \textbf{Concern-specific visual evidence:} The system combines a condition classifier for acne, pores, and wrinkles with supporting skin-type estimation and a CIOS module for count-informed ordinal acne-severity assessment. Each output remains within a non-diagnostic support role.

    \item \textbf{Deterministic grounding and risk controls:} The framework applies explicit checks to catalog evidence, privacy-sensitive data, unsafe or diagnostic-style language, approval-dependent actions, and response-specific evidence requirements.

    \item \textbf{Process-level auditability:} Structured traces and replay mechanisms record the route, tool use, verification outcomes, and response-finalization path. This extends assessment beyond the surface quality of the final answer.

    \item \textbf{Multi-axis evaluation:} The evaluation separates visual-model performance from agent behavior and examines visual reliability, tool use, grounding, safety and privacy behavior, failure handling, and process reproducibility.
\end{enumerate}

The contribution lies in the evaluated integration and control strategy. It does not depend on a claim that each vision model, retrieval component, or software module is individually novel.

The remainder of this paper is organized as follows. Section~2 reviews related work on visual skin analysis, acne-severity assessment, multimodal dermatology, agentic AI, and trustworthy user-facing systems. Section~3 presents the methodology, including visual routing, agent orchestration, grounding, verification, privacy controls, and trace design. Section~4 reports the visual-model and agent-process evaluations. Section~5 discusses the findings, practical implications, failure modes, and study limitations. Section~6 presents the conclusions and directions for further validation and deployment-oriented research.

\section{Related Work}

\subsection{Visual AI for Skin and Acne Analysis}

Artificial intelligence for facial skin analysis has expanded from broad image classification to consumer-oriented assessment of visible attributes and spatially localised facial features. Saiwaeo et al. treated normal, oily, and dry skin categories as a photographic classification problem and compared several convolutional backbones [\hyperlink{ref:19}{19}]. Their study showed that skin-type categories can be estimated from images, while also illustrating the limited scope of such labels. Other studies have moved beyond whole-image classification. Zheng et al. localised acne, pigmentation, and wrinkles in unconstrained selfies representing different skin tones, ages, severity levels, and lighting conditions [\hyperlink{ref:20}{20}]. Yoon et al. formulated wrinkles and pores as simultaneous morphological-segmentation targets, showing that different cosmetic concerns may require different spatial representations rather than a single general classifier [\hyperlink{ref:21}{21}]. These studies support multiattribute visual analysis, but an image-derived skin-type category should not be interpreted as an objective physiological measurement. Performance also remains sensitive to image acquisition and annotation conditions.

Acne analysis has developed a broader literature connecting lesion-level evidence with image-level severity. One smartphone-based system detected several acne-lesion types and supplied their counts to a separate severity grader, creating an explicit link between local detections and ordinal assessment [\hyperlink{ref:22}{22}]. A prior-knowledge-guided method instead segmented lesion regions and fused them with the original facial image to combine local evidence with global appearance [\hyperlink{ref:23}{23}]. Other studies have addressed uncertainty in the target labels. Label-distribution smoothing incorporates information about the grading scale into lesion-count learning [\hyperlink{ref:24}{24}], while AcneGrader uses ensemble pruning to combine selected deep models for image-level grading [\hyperlink{ref:25}{25}]. Detection-oriented research continues to improve local acne recognition with modern object detectors [\hyperlink{ref:26}{26}]. A recent study of self-supervised learning found that generic contrastive pretraining may remain poorly aligned with the subtle local features needed for fine-grained acne grading when labelled data are limited [\hyperlink{ref:27}{27}].

This literature also identifies several evaluation concerns that a single accuracy value cannot capture. Studies of facial features in unconstrained images report practical difficulties associated with lighting and skin-tone variation [\hyperlink{ref:20}{20}]. Experiments with limited acne labels show that performance can change substantially with the representation-learning procedure [\hyperlink{ref:27}{27}]. Rigorous validation therefore requires attention to duplicate and source control, identity-aware partitioning where possible, calibration, robustness testing, and evaluation on independent or source-held-out data. Even a well-validated image model remains a perception component. It does not determine when an acne-specific downstream tool should run, whether a product statement is supported by evidence, or whether a user-facing action is permitted.

\subsection{Multimodal and Foundation-Model Approaches in Dermatology}

Dermatology AI is moving beyond isolated image predictions toward workflows that combine visual information with contextual or language-based interaction. In teledermatology, an AI-assisted mole detector has been used to screen patient-uploaded images before specialist follow-up [\hyperlink{ref:28}{28}]. A broader review describes image models, large language models, and multimodal systems as related directions while identifying generalizability, bias, and explainability as unresolved concerns [\hyperlink{ref:29}{29}]. Research in cosmetic dermatology similarly presents AI as a means of supporting data-driven and personalised assessment while leaving interpretation and decision-making under human oversight [\hyperlink{ref:30}{30}]. These developments shift attention from image labelling alone to the controlled use and communication of visual evidence within a user-facing workflow.

Multimodal foundation models show that dermatological images can be integrated directly with conversational systems. SkinGPT-4 aligns a pretrained vision transformer with a large language model and was evaluated on real-world dermatology cases by board-certified dermatologists [\hyperlink{ref:31}{31}]. It provides an example of interactive reasoning over image and text inputs. Evaluation of multimodal LLMs on dermatoscopic images has nevertheless found substantial variation in reliability, including differences across sex and age groups [\hyperlink{ref:32}{32}]. A fluent multimodal interface is therefore insufficient evidence of reliable deployment. These systems are also oriented toward diagnosis, whereas the present work concerns non-diagnostic consumer support. That distinction affects the claims the system may make, the fallback behavior it requires, and the controls applied to recommendations and actions.

\subsection{LLM Agents, Tool Use, and Grounded Execution in Healthcare}

Agent-based systems extend language models beyond response generation by allowing them to decompose goals, coordinate roles, invoke tools, and interact with external state. MetaGPT implements this approach through standard operating procedures, role decomposition, structured intermediate artifacts, and verification of intermediate outputs [\hyperlink{ref:33}{33}]. Survey work commonly organises LLM-agent architectures around perception, planning or decision-making, and action in both multi-agent and human-agent settings [\hyperlink{ref:34}{34}]. A systematic review of LLM-based multi-agent systems identifies specialised collaboration, robustness, and trustworthy coordination as continuing research problems [\hyperlink{ref:35}{35}]. For a system such as SkinAgent, the relevant design requirements are explicit responsibilities, controlled state transitions, and verification around the LLM.

Grounded execution makes these responsibilities observable through the actions taken by the system. SAGE converts a user request into a sequence of discrete actions that can retrieve information, modify external state, verify execution, and determine when a task should end [\hyperlink{ref:36}{36}]. Evidence from healthcare tool use supports a similar conclusion. Goodell et al. compared retrieval, code interpretation, and task-specific calculation tools across large clinical-calculation experiments and found that machine-readable, task-specific tools produced the largest reductions in error [\hyperlink{ref:39}{39}].

These findings also show why the apparent quality of a final answer is an incomplete measure of a multistep agent. System reliability depends on whether the selected tools are appropriate, required tools are used, unnecessary or forbidden calls are avoided, arguments are correct, actions occur in a valid order, execution terminates properly, and tool failures are handled. These properties can be assessed through structured tool calls, state transitions, validators, and execution metadata without exposing private chain-of-thought reasoning.

External grounding is particularly important when a system recommends catalog items or reports product attributes. UniECS combines image, text, and multimodal retrieval through gated cross-modal fusion and evaluates these modes on a purpose-built e-commerce benchmark [\hyperlink{ref:37}{37}]. At an industrial scale, content-aware graph retrieval has been used to improve semantic representation and support long-tail and cold-start product search [\hyperlink{ref:38}{38}]. Such systems show how visual or textual intent can be linked to structured product records. Retrieval quality alone, however, does not establish the factual accuracy of every generated statement. In a skincare assistant, product identity, ingredients, price, stock, reviews, and other evidence-sensitive attributes should remain traceable to database fields or be marked as unavailable.

\subsection{Safety, Privacy, Explainability, and Human Oversight}

Health-adjacent generative systems require controls beyond predictive accuracy because errors may arise from the visual model, language model, tool sequence, or interactions among these components. Dermatology research identifies generalizability, bias, and explainability as persistent barriers to responsible deployment [\hyperlink{ref:29}{29}]. Evaluations of multimodal LLMs for skin-disease identification also show that reliability can vary across demographic groups [\hyperlink{ref:32}{32}]. These findings support targeted subgroup analysis rather than reliance on a single aggregate metric. Uncertainty handling, conservative fallback, escalation beyond the intended scope, and approval for action-sensitive operations should also be evaluated as system behaviors rather than described only as interface features.

Facial images introduce additional privacy and fairness concerns. PartialFace shows that a face-processing pipeline can transform visual information to make reconstruction more difficult while retaining recognition utility, illustrating that privacy preservation requires deliberate technical design [\hyperlink{ref:40}{40}]. The Casual Conversations dataset includes subject-level identifiers and annotations for age, gender, apparent skin tone, and lighting, allowing disparities to be examined across facial attributes and acquisition conditions [\hyperlink{ref:41}{41}]. These studies do not imply that another skincare system gains formal privacy or fairness guarantees by storing fewer images or reporting aggregate accuracy. Claims must instead be supported by system-specific tests of retention, access control, cross-user isolation, output leakage, and subgroup performance where suitable labels are available. When those labels are unavailable, the resulting uncertainty should be stated explicitly.

Human oversight also affects the strength of claims about safety and usefulness. The SMARTI protocol provides a prospective design in which AI-assisted dermatology is integrated into specialist practice, clinician decisions are compared before and after access to AI output, and safety is treated as an explicit endpoint [\hyperlink{ref:42}{42}]. SkinAgent has a different, non-diagnostic purpose, but the same methodological principle applies. Claims about safety, escalation quality, explanation usefulness, and recommendation appropriateness are stronger when evaluated by independent experts or users rather than inferred from developer-created examples alone. Explainability requires similar caution. Highlighting image regions or displaying intermediate evidence can improve inspectability, but it does not establish causal validity or clinical correctness.

\subsection{Synthesis and Research Gap}

The reviewed literature provides many of the components needed for a consumer-facing skincare system, but it does not yet establish a complete solution to the system-level problem considered here. Visual-analysis and teledermatology systems can localise visible concerns and support triage [\hyperlink{ref:20}{20}], [\hyperlink{ref:28}{28}]. Multimodal dermatology models can connect images with conversational reasoning [\hyperlink{ref:31}{31}]. Agent frameworks can define roles and grounded action sequences [\hyperlink{ref:33}{33}], [\hyperlink{ref:36}{36}], while e-commerce retrieval systems can link visual or textual intent to structured product records [\hyperlink{ref:37}{37}]. Healthcare tool-use studies show the value of task-specific external computation [\hyperlink{ref:39}{39}]. Privacy and prospective-evaluation research further shows that risk controls and human assessment require separate evaluation [\hyperlink{ref:40}{40}], [\hyperlink{ref:42}{42}].

What remains insufficiently studied is an auditable, non- \linebreak diagnostic skincare framework that combines visual concern routing, condition-dependent acne-severity analysis, bounded LLM planning, typed tools, database-grounded product evidence, deterministic safety and privacy checks, approval-controlled actions, response-specific evidence requirements, and replayable execution traces. These components must be assessed under a unified protocol that examines visual reliability, tool selection, action validity, grounding, safety and privacy behavior, failure recovery, and trace reproducibility. Independent expert review and external validation should also be included when available. The research gap lies in the integration and evaluation of these components rather than in the individual novelty of the classifier, detector, database, retrieval model, or LLM.

\section{Methods}

\subsection{Study Design, Intended Use, and Claim Boundary}

SkinAgent AI was developed as a research prototype for non-diagnostic skincare support. The system combines dedicated visual models, a primary LLM-based conversational orchestrator, typed backend tools and application programming interfaces (APIs), a database-backed product layer, deterministic safety and privacy controls, report and cart workflows, and structured execution traces. Its intended functions include skincare education, summarization of visible concerns, evidence-bounded product assistance, and low-risk application actions. The system is not intended to diagnose a medical condition, prescribe medication, provide a treatment plan, or replace a dermatologist or other licensed clinician. Visual outputs are therefore described as model-estimated attributes or support signals rather than clinical findings.

The evaluation is divided into model-level and system-level components. Model-level evaluation covers the visual modules for skin-type estimation, skin-condition routing, and count-informed acne-severity support. System-level evaluation examines the orchestrator and its surrounding controls, including intent routing, tool use, evidence grounding, safety and privacy checks, fallback behavior, and trace replay. This distinction prevents image-classification accuracy from being treated as evidence of reliable agent behavior. It also prevents successful software orchestration from being interpreted as clinical validation of the visual models.

The evaluated architecture separates end-user content from technical telemetry. Patient View presents concise, non-
\linebreak diagnostic guidance and excludes internal identifiers, debugging fields, private payloads, and execution telemetry. Expert/Admin views may present controlled research metadata, including model-estimated confidence returned by an actual service, grounding and safety states, trace summaries, and reliability information. Hidden prompts, private session tokens, raw image payloads, and private model reasoning are excluded from both views.

The study materials include an internal development and regression suite containing 189 scripted agent cases and 30 rubric cases. They also include a locked but non-independent 240-case system benchmark and a frozen 60-case controlled comparison of configurations C and D. These evaluations were conducted within the research environment and are not presented as independent external or clinical validation.

\subsection{SkinAgent System Architecture and Execution Lifecycle}

The architecture follows a bounded Plan--Execute--Verify lifecycle. An authenticated request enters the backend with the conversation state, user preferences, optional image metadata, and client request identifiers. Request-level controls first reject unsupported, privacy-invasive, or out-of-scope content. The intent router then assigns the request to an operational category, such as visual analysis, product recommendation, product comparison, cart action, report generation or retrieval, privacy information, search, greeting, or fallback. Simple requests may follow deterministic routes, while more complex requests can be passed to the LLM-based planner to construct an execution path over an approved tool registry.

The architecture assigns separate responsibilities to probabilistic and deterministic components. The LLM interprets natural-language requests, plans when necessary, synthesizes user-facing explanations, and may critique or revise a draft response. Deterministic components enforce tool schemas, permissions, privacy boundaries, evidence requirements, non-diagnostic wording rules, and action-approval requirements. Dedicated tool interfaces invoke visual models and external or local visual services. Product, ingredient, report, cart, and retrieval functions are also exposed as bounded tools rather than simulated by the language model. Most supporting components are therefore deterministic guards, service wrappers, retrieval modules, or offline evaluators rather than autonomous agents.

After tool execution, evidence-assembly modules collect the available visual summaries, product records, report identifiers, safety profiles, and response-mode requirements. Deterministic checks then examine safety, privacy, product grounding, tool use, and evidence completeness. If required evidence is unavailable or the draft exceeds the configured claim boundary, the system can request a revision or return a conservative fallback response.

\begin{figure*}[!t]
\centering
\includegraphics[width=0.98\textwidth,keepaspectratio]{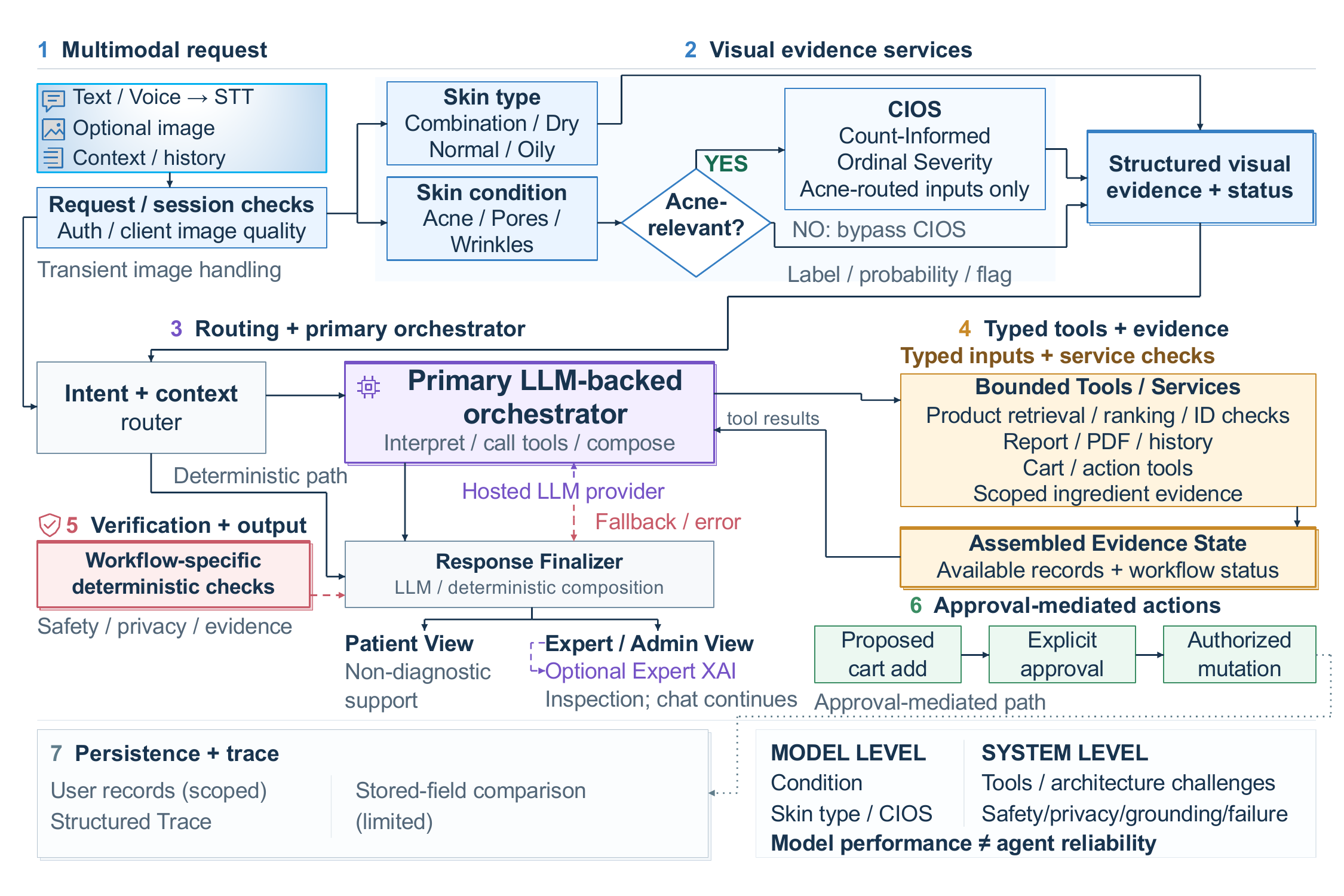}
\caption{SkinAgent AI architecture and bounded execution workflow. Multimodal text, voice, image, and contextual inputs pass through request and session checks before visual outputs are normalized into structured, non-diagnostic evidence. Skin-type estimation precedes skin-condition routing in the implemented workflow, and the Count-Informed Ordinal Severity (CIOS) module is invoked only for acne-relevant inputs. A primary LLM-based orchestrator coordinates bounded tools and services, while workflow-specific deterministic checks constrain safety, privacy, evidence use, and response finalization. State-changing actions require explicit user approval, and structured execution traces support auditing and reliability analysis. Model-level visual evaluation and system-level agent evaluation are conducted separately.}
    \label{fig:architecture}
\end{figure*}

The finalization stage produces Patient View content, cards or attachments, and controlled Expert/Admin metadata. Conversation records, reports, pending actions, grounding records, and privacy-filtered execution traces may then be stored for reliability analysis. Auditability is based on logged routing decisions, tool calls, verification outcomes, and replayable process states rather than disclosure of hidden chain-of-thought tokens. Figure~\ref{fig:architecture} summarizes the architecture and execution lifecycle.

\pdfbookmark[2]{Figure 1}{bm-figure-1}

\subsection{Visual Input Validation and Image-Handling Boundary}

Facial images pass through an engineering-quality gate before visual inference. The gate evaluates image resolution, brightness or luminance, contrast, blur, face coverage, face centeredness, and face count. It assigns one of three outcomes: PASS, WARNING, or FAIL. PASS permits normal visual processing. WARNING permits processing with a quality caveat and more conservative language. FAIL blocks analysis and asks the user to capture another image. The architecture requires one usable face; images containing no face or multiple faces therefore receive a FAIL outcome. These controls are intended to reduce unsupported inference from unsuitable images. The study materials do not report benchmark-level sensitivity, specificity, false-acceptance rate, or false-rejection rate for the quality gate.

Image handling follows a session-limited privacy policy. Raw uploaded image bytes, base64 encodings, browser object or Blob URLs, facial crops, and local file paths are designated as non-persistent. According to the documented implementation, these data are not stored in MongoDB, reports, browser local storage, or audit logs. Temporary browser-preview URLs are revoked after use. Persisted records contain only the structured summaries required for conversation continuity, reports, grounding, or evaluation.

Optional visual-explanation overlays are treated as session-only Expert View data and are not stored. If a visual-analysis service, explanation service, or lesion-localization output is unavailable, the system must describe the evidence as unavailable or uncertain. It must not fabricate lesion coordinates, heatmaps, confidence values, or other visual evidence.

\subsection{Skin-Type Estimation}

The supporting skin-type module estimates one of four visual categories: Combination, Dry, Normal, or Oily. The skin-type experiments used the publicly accessible skin type dataset from Roboflow Universe [\hyperlink{ref:44}{44}], comprising 14,181 images in the study release. These images are divided into 12,408 training images, 1,181 validation images, and 592 test images.

The selected classifier is EfficientNetV2-M, implemented in PyTorch with the \texttt{timm} model library and initialized with ImageNet-pretrained weights. Images are resized to $384\times384$ pixels and normalized using the ImageNet mean and standard deviation.

Training uses AdamW with weight decay, a warm-up period followed by cosine learning-rate decay, gradient accumulation, and an exponential moving average of the model parameters. Data augmentation and regularization include random resized cropping, horizontal flipping, RandAugment, color augmentation, MixUp, CutMix, and label smoothing.

During inference, the network returns a probability distribution over the four categories. The category with the highest probability is converted into a structured, model-estimated skin-type attribute. This attribute provides personalization context for downstream recommendation logic and is not interpreted as a physiological measurement or clinical diagnosis. Performance outcomes are reported in the Results section.

\subsection{Skin-Condition Classification and Routing}

The skin-condition task contains three classes: Acne, Pores, and Wrinkles. The three-class study dataset was constructed from the Acne, Pores, and Wrinkles categories of the publicly accessible Skin Issues Dataset [\hyperlink{ref:45}{45}]. After the documented group-safe duplicate-control procedure, the final study dataset contained 5,394 images divided into 3,775 training images, 809 validation images, and 810 test images. The training set contained 1,359 acne images, 1,044 pore images, and 1,372 wrinkle images. The validation set contained 291, 224, and 294 images, respectively, while the test set contained 292, 224, and 294 images.

Evaluation used the documented group-safe, perceptual-duplicate-controlled partition identified as \texttt{Dataset\_GroupSafe\_dhash5}. The study materials document dHash-based perceptual-duplicate grouping. The final canonical split manifest was not retained in the release package, which limits independent reconstruction of the exact test-set membership.

Four backbone families are evaluated under the same three-class protocol: EfficientNetV2-S, ConvNeXt-Tiny, DeiT-S, and DINOv2 ViT-S/14. EfficientNetV2-S, ConvNeXt-Tiny, and DeiT-S use $384\times384$ inputs, while DINOv2 ViT-S/14 uses $392\times392$ inputs. Training uses transfer learning, AdamW, weighted cross-entropy with label smoothing, mixed-precision computation, cosine learning-rate scheduling, and early stopping based on validation macro-F1. The checkpoint with the highest validation macro-F1 is selected. Augmentation consists of random resized cropping, horizontal flipping, mild color jitter, and small-angle rotation. The selected EfficientNetV2-S configuration is repeated with seeds 42, 123, and 2025 using the same group-safe partitions.

The evaluation metrics are accuracy, balanced accuracy, macro-F1, weighted F1, macro-precision, macro-recall, one-vs-rest macro-AUROC, macro-AUPRC, expected calibration error (ECE), Brier score, acne miss rate, and non-acne false-routing rate. Acne miss rate is the proportion of acne images classified as pores or wrinkles. Non-acne false-routing rate is the proportion of pore or wrinkle images incorrectly routed to the acne branch.

Formally, let \(y_i\) denote the reference class and \(\hat{y}_i\) the predicted class for image \(i\). The acne miss rate (AMR) is defined as

\begin{equation}
\mathrm{AMR}
=
\frac{
\left|\left\{i:y_i=\mathrm{Acne},\ \hat{y}_i\neq\mathrm{Acne}\right\}\right|
}{
\left|\left\{i:y_i=\mathrm{Acne}\right\}\right|
}.
\label{eq:amr}
\end{equation}

The non-acne false-routing rate (NAFR) is defined as

\begin{equation}
\mathrm{NAFR}
=
\frac{
\left|\left\{i:y_i\in\{\mathrm{Pores},\mathrm{Wrinkles}\},\
\hat{y}_i=\mathrm{Acne}\right\}\right|
}{
\left|\left\{i:y_i\in\{\mathrm{Pores},\mathrm{Wrinkles}\}\right\}\right|
}.
\label{eq:nafr}
\end{equation}

Model selection therefore considers both overall classification quality and the operational consequences of downstream routing errors.

Probability calibration is evaluated because routing may depend on confidence information. The supplied materials do not document a fixed probability threshold for acne routing. No threshold is introduced in this study beyond the documented class and routing outputs.

Within the complete workflow, the condition output is normalized into structured evidence before reaching the conversational layer. An acne-relevant result enables the CIOS module. A non-acne result remains available for general supportive guidance and product-context reasoning without forcing acne-specific analysis. The classifier therefore functions as a routing component rather than a standalone endpoint. Calibrated probabilities and class-specific routing errors are retained because the downstream effect of missing an acne case differs from that of temporarily routing a non-acne case to the acne branch. This routing policy is not presented as clinical triage.

\subsection{Count-Informed Ordinal Acne-Severity Support}

The Count-Informed Ordinal Severity module provides ordinal acne-severity support for acne-relevant inputs. It uses the ACNE04 dataset [\hyperlink{ref:4}{4}] while preserving the original lesion-detection partitions: 1,049 training images, 116 validation images, and 292 test images. The four operational categories are Mild, Moderate, Severe, and Very Severe.

Image-level reference categories are generated from lesion annotations using the documented count rule: $0\text{--}4$ lesions for Mild, $5\text{--}19$ for Moderate, $20\text{--}49$ for Severe, and at least 50 lesions for Very Severe. This construction is reported explicitly because the supplied materials do not describe these targets as independently assigned dermatologist severity grades.

For acne-routed inputs, SkinAgent invokes the fixed CIOS module. The module combines an image-level severity estimate with lesion-count information returned by a fixed visual-counting component and produces one of the four ordinal support categories. CIOS is treated as a supporting visual service within SkinAgent rather than as a separately proposed acne-grading architecture. Its internal representation and component-integration procedure are not presented as methodological contributions of SkinAgent.

The CIOS output is used only as bounded visual evidence for downstream, non-diagnostic support. It can inform conservative messaging and escalation logic but is not described as a clinical diagnosis. Evaluation uses classification metrics and quadratic weighted kappa (QWK) to account for the ordered relationship among the four categories. The evaluation status and reproducibility limitations of CIOS are reported in the Results section.

\subsection{LLM Orchestration, Tool Use, and Database Grounding}

The conversational layer uses one primary LLM-based orchestrator supported by deterministic modules and typed tools. Intent routing determines whether a request concerns visual analysis, product recommendation, product comparison, cart operation, report generation or retrieval, privacy, search, greeting, or fallback. For requests that require several services, the planner constructs a bounded execution path. The typed tool registry exposes only approved operations, and the tool-use verifier checks schemas, permissions, arguments, user scope, and privacy boundaries before execution.
The archived study materials do not preserve the exact hosted LLM model identifier, model version or alias, decoding temperature, or top-\(p\) setting used during the final benchmark runs. This limits exact replay of LLM-mediated outputs and is treated as a reproducibility limitation.

Tool and API modules provide access to visual services, the product database, ingredient or guideline evidence, report services, cart operations, and other configured retrieval functions. Product recommendations are grounded in the database. Candidate products are retrieved from the catalog, filtered according to the active safety and context profile, and validated against stored identifiers such as product IDs or slugs. Product comparisons resolve explicit records or previously recommended items and compare only available database fields. Review summaries use stored review data when available and disclose missing evidence rather than inventing customer feedback. Product names, attributes, prices, stock status, ratings, ingredients, and review claims must therefore be derived from retrievable records whenever they appear in a response.

Actions that change system state are separated from informational generation. A general request to add recommended products does not immediately modify the cart. The system first creates pending actions and requires explicit user approval before updating the cart. Report generation similarly identifies the latest usable analysis, selects the appropriate report mode, creates a report identifier, and provides open or download actions.

Service failures trigger conservative fallback behavior. Unavailable visual or retrieval evidence is represented as unavailable or uncertain rather than replaced by language-model fabrication. Finalization then produces a user-facing response and controlled metadata that are consistent with the active response mode.

\subsection{Safety, Privacy, Evidence, and Approval Controls}

Safety controls operate throughout the execution lifecycle rather than through a disclaimer alone. Request-level guards and image-quality checks run before tool execution. Safety triage and cross-modal disagreement handling operate on the assembled evidence. Deterministic critic gates inspect the draft response for diagnostic claims, prescriptions, dosage instructions, cure claims, treatment plans, dermatologist-equivalent language, and unsupported certainty.

High-risk or uncertain states can suppress targeted product guidance, withhold unsupported severity or confidence statements, request a replacement image, or shift the response toward consultation-first language. Product safety profiles are applied before recommendation when the relevant information is encoded. These controls define a bounded, non-diagnostic claim surface. They do not provide a formal guarantee of clinical safety.

Privacy controls enforce session-limited image handling and output filtering. Under the documented policy, raw images, base64 data, object URLs, facial crops, local paths, API keys, hidden prompts, session tokens, and private model reasoning are excluded from persistent records and user-facing responses. User-owned resources are accessed through authenticated, user-scoped records, while role-based controls restrict administrative and research interfaces. Patient View omits debugging identifiers and guard telemetry. Expert/Admin views may expose controlled summaries but do not reveal private payloads or hidden reasoning. The system is therefore described as privacy-aware and data-minimizing rather than formally privacy-preserving.

Evidence requirements depend on the response mode. Analysis, fallback, high-risk, product, cart, report, privacy, and out-of-scope responses do not require the same fields. Intentional suppression of unsupported information is therefore not treated as a missing-data error. Product-grounding checks validate database-backed statements, and the tool-use verifier blocks forbidden or privacy-unsafe operations. Approval is mandatory before cart mutation. If required evidence or services are unavailable, the finalizer returns a conservative fallback or states that the evidence is unavailable.

This evidence contract is deliberately asymmetric. A routine product response requires catalog-backed product evidence but does not need to expose internal model telemetry. A high-risk or uncertain response may intentionally omit product suggestions or severity confidence. A privacy response does not require visual-analysis evidence. A cart action requires a resolvable product and explicit approval. These distinctions prevent a generic completeness rule from encouraging the system to invent fields solely to satisfy a schema. They also provide a measurable difference between an intentionally conservative response and an incomplete execution.

\begin{figure*}[!t]
\centering
\includegraphics[width=0.95\textwidth]{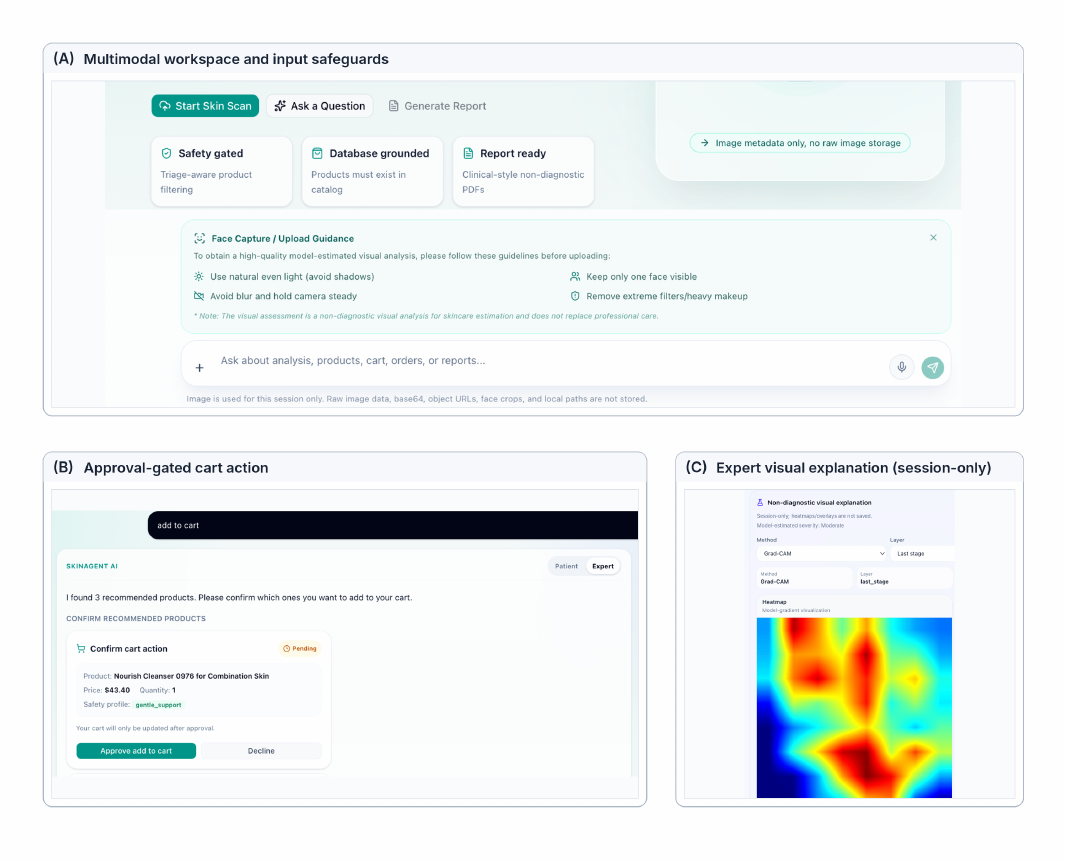}
\caption{Representative SkinAgent interfaces and controls. (A) The multimodal workspace provides non-diagnostic image-analysis entry points, image-quality guidance, database-grounded support, and session-limited image handling. (B) State-changing cart operations are represented as pending actions and require explicit user approval before execution. (C) Expert View provides an optional, session-only visual explanation for model inspection. The visualization is not persisted and is not interpreted as clinical localisation evidence.}
\label{fig:interfaces}
\end{figure*}

Figure~\ref{fig:interfaces} shows representative interfaces for multimodal input, an approval-gated cart action, and the optional session-only Expert visual-explanation view. A complete demonstration of the implemented SkinAgent AI prototype is available in the accompanying system walkthrough [\hyperlink{ref:46}{46}].

\pdfbookmark[2]{Figure 2}{bm-figure-2}

\subsection{Agent Benchmark, Baselines, and Ablation Design}

The internal development and regression evaluation contains 189 scripted agent cases and 30 rubric cases. The scripted suite covers intent routing, tool selection, forbidden-tool checks, safety and privacy behavior, evidence completeness, adversarial and prompt-injection inputs, cross-modal disagreement, product and comparison requests, cart confirmation, report workflows, image warnings and fallbacks, recovery behavior, and trace replay.

The rubric covers twelve communication dimensions: adherence to the non-diagnostic boundary, escalation, context awareness, image-quality handling, uncertainty communication, product grounding, avoidance of clinical claims, clarity, clarification behavior, privacy safety, Patient View cleanliness, and Expert View audit metadata. Because these cases informed development, they are treated as internal reliability evidence rather than an independent test set.

The final system evaluation also includes a locked but non-independent 240-case benchmark and a frozen 60-case controlled comparison of configurations C and D. The 240-case benchmark uses prespecified requests and expected behaviors. It measures intent accuracy, exact tool-set match, required-tool recall, forbidden-tool use, task completion, safety, privacy and grounding violations, fallback and escalation correctness, and trace-level behavior.

Configuration C consists of a tool-using LLM with database grounding and minimal deterministic controls. Configuration D is the complete SkinAgent system, incorporating planning, typed tools, grounding, safety, privacy and evidence verification, approval checks, critic and fallback behavior, and trace logging. Direct-LLM configuration A and safety-prompt-only configuration B were retained as prospective baselines but were not completed as manuscript-ready quantitative comparisons. The reported architectural analysis is therefore limited to configurations C and D.

Tool use is scored at several levels rather than through exact-set matching alone. Exact tool-set match requires the observed and expected tool sets to be identical. Required-tool recall measures whether all task-critical tools were invoked. Extraneous-tool rate measures unnecessary calls, while forbidden-tool violation rate records disallowed operations. Argument and schema validity assess whether typed arguments satisfy the tool contract and user scope. Sequence validity examines ordering requirements, such as recommendation before confirmation and approval before cart mutation. Task and action completion record whether the requested operation was completed under the applicable policy.

For benchmark case \(i\), let \(E_i\) denote the expected tool set, \(O_i\) the observed tool set, and \(F_i\) the set of tools forbidden for that case. Across \(N\) benchmark cases, exact tool-set match is

\begin{equation}
\mathrm{ETM}
=
\frac{1}{N}
\sum_{i=1}^{N}
\mathbf{1}\!\left(O_i=E_i\right),
\label{eq:etm}
\end{equation}

where \(\mathbf{1}(\cdot)\) is an indicator function that equals 1 when the condition is true and 0 otherwise. The forbidden-tool violation rate is

\begin{equation}
\mathrm{FVR}
=
\frac{1}{N}
\sum_{i=1}^{N}
\mathbf{1}\!\left(O_i\cap F_i\neq\varnothing\right).
\label{eq:fvr}
\end{equation}

Let \(c_i=1\) when benchmark case \(i\) satisfies the predefined strict completion criterion under the applicable policy and \(c_i=0\) otherwise. Strict task completion (STC) is therefore

\begin{equation}
\mathrm{STC}
=
\frac{1}{N}
\sum_{i=1}^{N} c_i.
\label{eq:stc}
\end{equation}

Outcome values are reported in the Results section.

The 240-case benchmark was frozen before final scoring but remains non-independent because it was conducted within the research environment and was not authored or reviewed independently of system development. The 60-case comparison was also frozen before execution and was subsequently assessed through structured post-hoc adjudication against case-specific expected behaviors. Neither evaluation is treated as clinical or independent external validation.

\subsection{Robustness, Failure-Mode, and Human/Expert Evaluation}

The completed project documentation contains manual and scripted failure-oriented tests rather than an external robustness study. Manual tests cover authentication, facial-image upload, gatekeeper PASS/WARNING/FAIL behavior, visual-service fallback, report and PDF generation, explanation-service availability, product recommendations and ingredient evidence, cart and checkout workflows, cross-user isolation, administrative diagnostics, and privacy-audit queries.

The scripted evaluation additionally includes adversarial and prompt-injection cases, privacy-leakage cases, cross-modal disagreement, image warnings and fallbacks, and recovery scenarios. If a visual API is unavailable, the specified behavior is an UNCERTAIN or conservative fallback. If visual-explanation output is unavailable, the main conversation must continue without fabricating an overlay. An empty or unavailable product catalog must not be replaced with invented product evidence.

The architecture also includes a MongoDB privacy audit that searches for prohibited persistent fields associated with raw images, base64 data, object URLs, facial crops, or local file paths. Failure handling is assessed at the process level. A service failure is handled correctly only when the system records an unavailable or uncertain state, preserves authorization and data boundaries, and avoids fabricated evidence or duplicate side effects. Cart workflows require confirmation during retries, and failures in report or visual services must not be converted into synthetic analysis results. These tests assess implementation behavior and do not establish formal privacy guarantees.

The supplied materials do not document completed controlled robustness experiments for blur, brightness, compression, occlusion, or source-held-out visual generalization. They also do not document a completed independent dermatologist or domain-expert evaluation. No human or expert outcome is therefore reported as part of the completed study. Latency measurements at the 50th and 95th percentiles are also unavailable. These evaluations should be added only if they are completed under a frozen protocol before submission.

\subsection{Statistical Analysis and Reproducibility}

The skin-condition experiments report conventional multiclass measures together with calibration and routing-sensitive metrics. The selected architecture is evaluated across three independent training seeds, 42, 123, and 2025, using the same group-safe partitions. Repeated-run performance is summarized using the mean and sample standard deviation.

The skin-condition protocol reports accuracy, balanced accuracy, macro-F1, weighted F1, macro-precision, macro-recall, one-vs-rest macro-AUROC, macro-AUPRC, ECE, Brier score, acne miss rate, and non-acne false-routing rate. The acne-severity protocol reports accuracy, macro-F1, class-level measures, confusion analysis, and QWK. For proportion-based system outcomes, 95\% Wilson confidence intervals are reported where applicable. Zero-event safety, privacy, and approval outcomes are accompanied by one-sided exact 95\% upper confidence bounds. Skin-type macro-F1 is reported with a 10,000-resample bootstrap 95\% confidence interval.

Paired binary outcomes for configurations C and D are compared using exact two-sided McNemar tests on the same frozen cases. The three-seed skin-condition summaries are descriptive means and sample standard deviations rather than inferential significance tests.

Reproducibility depends on preserving the dataset partitions, training seeds, checkpoint-selection rules, and documented preprocessing and training settings. The architecture also maintains privacy-filtered execution traces containing routed intent, planned and executed tools, schema checks, safety decisions, evidence states, and the finalization path. Raw facial images and hidden model reasoning are excluded from these traces. Trace replay is intended to assess process concordance without retaining prohibited private payloads.

The study archive contains the available dataset partitions, training seeds, evaluation scripts, benchmark definitions,\linebreak statistical-analysis files, and privacy-filtered execution traces used in the analysis. Exact bitwise reproduction of outputs from hosted LLMs is not expected because provider-side models and APIs may change over time. Reproducibility is therefore assessed through the documented configuration, benchmark inputs, tool traces, and aggregate results, subject to the artifact limitations described above.

\subsection{Ethical Considerations}

This study did not recruit human participants, conduct clinical interventions, or collect new identifiable participant data. The visual experiments used pre-existing, publicly accessible datasets: ACNE04 [\hyperlink{ref:4}{4}], the Roboflow skin-type dataset [\hyperlink{ref:44}{44}], and the Kaggle skin-condition dataset [\hyperlink{ref:45}{45}]. No attempt was made to identify or re-identify individuals represented in these datasets.

Formal institutional ethics review was not sought because the study was limited to secondary analysis of existing datasets and scripted software and system evaluations, with no prospective participant interaction or new data collection. No additional informed consent was obtained for the present analysis.

\section{Results}

\subsection{Skin-Condition Routing Performance}

The skin-condition experiment evaluated Acne, Pores, and Wrinkles using a group-safe, duplicate-controlled dataset of 5,394 images. The partition contained 3,775 training images, 809 validation images, and 810 test images. Four backbone families were evaluated under the same three-class protocol: EfficientNetV2-S, ConvNeXt-Tiny, DeiT-S, and DINOv2 ViT-S/14. EfficientNetV2-S was selected for the routing module. The release package did not retain a traceable numerical comparison of all four backbones. Comparative backbone results are therefore not reconstructed, and quantitative reporting is limited to the selected model's three-seed summary.

Across seeds 42, 123, and 2025, EfficientNetV2-S achieved a mean accuracy of $99.84\%\pm0.07\%$ and a macro-F1 of $99.85\%\pm0.07\%$. The mean macro-AUROC was $99.9984\%\pm0.0026\%$, and the mean macro-AUPRC was $99.9972\%\pm0.0046\%$. Expected calibration error was $0.00975\pm0.00137$.

Routing-specific error rates were also low. The mean acne miss rate was $0.23\%\pm0.20\%$, while the mean non-acne false-routing rate was $0.13\%\pm0.11\%$. An acne miss prevents the acne-specific severity module from being invoked. A false acne-routing event sends an image of pores or wrinkles to the acne-specific branch. Both error rates remained below 0.5\% on average. No additional threshold-specific operating point was retained in the release package, so these values apply only to the documented classification and routing outputs.

The three-seed results showed little variation across runs on the fixed internal test set. They are presented as a stability analysis rather than an inferential test of superiority. No independent external condition dataset or completed source-held-out evaluation was included. Class-level confusion estimates and confidence intervals were also unavailable in the traceable release package. TThe release package did not retain the final 810-image test manifest, per-seed probability files, or checkpoint hashes required for independent recomputation; no missing values were reconstructed.

\pdfbookmark[2]{Table 1}{bm-table-1}
\begin{table*}[!t]
\centering
\small
\caption{Skin-condition routing performance and three-seed stability.}
\label{tab:res1}
\begin{tabularx}{\textwidth}{|X|X|X|X|}
\hline
\textbf{Model} & \textbf{Test set} & \textbf{Metric} & \textbf{Three-seed result} \\
\hline
EfficientNetV2-S & 810 images & Accuracy & $99.84\%\pm0.07\%$ \\
\hline
EfficientNetV2-S & 810 images & Macro-F1 & $99.85\%\pm0.07\%$ \\
\hline
EfficientNetV2-S & 810 images & Macro-AUROC & $99.9984\%\pm0.0026\%$ \\
\hline
EfficientNetV2-S & 810 images & Macro-AUPRC & $99.9972\%\pm0.0046\%$ \\
\hline
EfficientNetV2-S & 810 images & ECE & $0.00975\pm0.00137$ \\
\hline
EfficientNetV2-S & 810 images & Acne miss rate & $0.23\%\pm0.20\%$ \\
\hline
EfficientNetV2-S & 810 images & False acne-routing rate & $0.13\%\pm0.11\%$ \\
\hline
\end{tabularx}
\end{table*}

\noindent\textit{Note.} Values are the documented three-seed summary. The release package did not retain the per-seed probability files or checkpoint hashes required for independent recomputation.

\subsection{Supporting Visual Subsystems}

The skin-type module was evaluated on a frozen test set of 592 images using EfficientNetV2-M. It correctly classified 526 images, corresponding to an accuracy of 88.85\% (95\% Wilson confidence interval [CI]: 86.06\%--91.14\%). Macro-F1 was 0.8924, with a 10,000-resample bootstrap 95\% CI of 0.8660--0.9168. These values were reproduced from the retained prediction file and metric script. The output is interpreted as a model-estimated visual skin-type category rather than a clinical or physiological diagnosis.

The retained prediction file allowed the test denominator and confidence intervals to be reconstructed directly. The analysis remains limited to the prespecified accuracy and macro-F1 outcomes. The confidence intervals quantify sampling uncertainty on the frozen internal test set and do not extend the claim beyond the visual-category classification task.

The CIOS module was retained as an integrated support service for acne-routed inputs. Aggregate evaluation results were available for the canonical ACNE04 test split. However, the final per-sample ensemble predictions and complete checkpoint provenance needed for strict independent recomputation were not retained. Standalone CIOS performance is therefore reported as aggregate internal evaluation evidence rather than as an independently reproducible or externally validated outcome. Instead, CIOS is reported as a bounded, count-informed ordinal support component within the SkinAgent workflow rather than an independently validated clinical severity-grading system.

\pdfbookmark[2]{Table 2}{bm-table-2}
\begin{table*}[!t]
\centering
\small
\caption{Performance of the supporting visual subsystems.}
\label{tab:res2}
\begin{tabularx}{\textwidth}{|X|X|X|X|X|X|}
\hline
\textbf{Subsystem} &
\textbf{Model/task} &
\textbf{$N$} &
\textbf{Accuracy} &
\textbf{Macro-F1} &
\textbf{Additional metric} \\
\hline
Skin type &
EfficientNetV2-M &
592 &
88.85\% &
0.8924 &
Accuracy 95\% CI: 86.06\%--91.14\%; F1 95\% CI: 0.8660--0.9168 \\
\hline
Acne severity &
CIOS module &
292 &
84.59\% &
0.8149 &
QWK = 0.9076 \\
\hline
\end{tabularx}
\end{table*}

\noindent\textit{Note.} Skin-type results were reproduced from the retained per-sample predictions. Acne-severity results were reproduced from the recovered aggregate confusion matrix; the final per-sample severity predictions were not retained.

\subsection{Agentic Orchestration and Tool-Use Performance}

The locked system benchmark contained 240 cases and was treated as a non-independent evaluation. Intent accuracy was 80.00\% (192/240; 95\% Wilson CI: 74.48\%--84.57\%). Exact tool-set match was 62.92\% (151/240; 95\% CI: 56.65\%--68.78\%), and mean per-case required-tool recall was 66.25\%. Strict task completion was 47.08\% (113/240; 95\% CI: 40.86\%--53.39\%). Eight forbidden-tool violations occurred, corresponding to 3.33\% of cases (95\% CI: 1.70\%--6.44\%).

These results distinguish request recognition from execution. Correctly identifying the intent did not ensure selection of the complete tool set or successful completion of the requested action. Exact tool-set match and required-tool recall were reported separately because a trajectory could invoke a required tool while still omitting another required tool or adding an unnecessary one. The eight forbidden-tool events were retained as a separate outcome rather than absorbed into a single success score.

\pdfbookmark[2]{Table 3}{bm-table-3}
\begin{table*}[!t]
\centering
\small
\caption{Agentic orchestration and tool-use performance.}
\label{tab:res3}
\begin{tabularx}{\textwidth}{|X|X|X|X|X|}
\hline
\textbf{Evaluation} &
\textbf{$N$} &
\textbf{Metric} &
\textbf{Result} &
\textbf{95\% interval/note} \\
\hline
Locked benchmark &
240 &
Intent accuracy &
80.00\% (192/240) &
74.48\%--84.57\% \\
\hline
Locked benchmark &
240 &
Exact tool-set match &
62.92\% (151/240) &
56.65\%--68.78\% \\
\hline
Locked benchmark &
240 &
Mean required-tool recall &
66.25\% &
Mean per-case recall \\
\hline
Locked benchmark &
240 &
Strict task completion &
47.08\% (113/240) &
40.86\%--53.39\% \\
\hline
Locked benchmark &
240 &
Forbidden-tool violations &
3.33\% (8/240) &
1.70\%--6.44\% \\
\hline
Development suite &
189 &
Required-tool recall &
55.88\% (57/102) &
Internal scripted evidence \\
\hline
Rubric suite &
30 &
Mean rubric score &
95.5/100 &
Internal scripted evidence \\
\hline
\end{tabularx}
\end{table*}

\pdfbookmark[2]{Figure 3}{bm-figure-3}
\begin{figure*}[!t]
\centering
\includegraphics[width=0.95\textwidth]{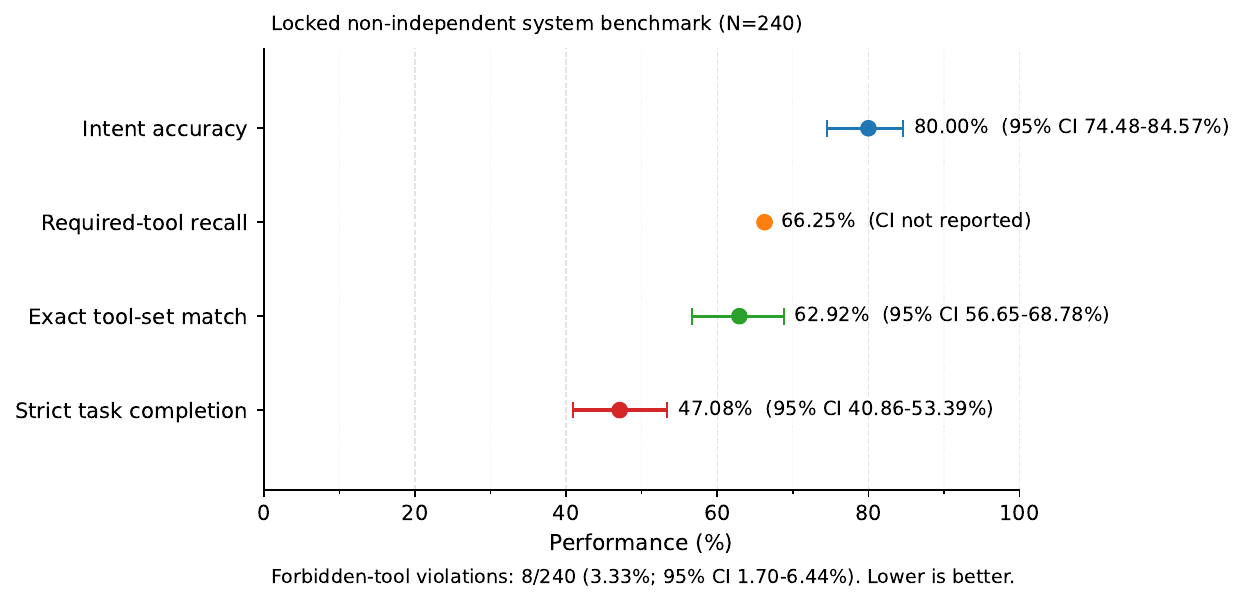}
\caption{Agentic orchestration performance on the locked, non-independent 240-case benchmark. Point estimates are shown for intent accuracy, required-tool recall, exact tool-set match, and strict task completion. Ninety-five percent confidence intervals are included where available. Required-tool recall is shown without a manuscript-facing confidence interval because none was reported in the final analysis. Forbidden-tool violations are reported separately because lower values indicate better performance.}
\label{fig:locked-benchmark}
\end{figure*}

Under the locked benchmark definitions, no safety, privacy, or grounding violation was recorded in the 240 cases. Each outcome was 0/240, with a one-sided exact 95\% upper confidence bound of 1.24\%. Fallback and escalation correctness were each 240/240, with a 95\% Wilson lower bound of 98.42\%. These finite-sample observations do not establish universal safety, privacy, grounding, fallback, or escalation performance. Final aggregate schema-validity and sequence-validity results were not retained in the release package and are therefore not reported.

Perfect fallback and escalation results did not offset the task-completion and tool-selection errors recorded in the same benchmark. The evaluation is therefore reported as a multidimensional system assessment rather than reduced to a single success rate.

The earlier development and regression suite contained 189 scripted cases. Expected intent was injected into the controller, so the 189/189 result represents expected-intent/controller concordance rather than independent intent-classifier accuracy. A metric-definition audit identified 57 correctly invoked required-tool slots among 102 required slots, corresponding to a micro-averaged required-tool recall of 55.88\%. Raw evidence completeness was 91\%. Mode-adjusted evidence completeness, trace-replay concordance, safe-fallback completeness, high-risk escalation completeness, and product-grounding completeness were each 100\% within this internal scripted suite.

A separate 30-case internal rubric evaluation produced a mean score of 95.5/100. This value corresponds to the rounded 95/100 result used in the earlier study record.

The development and locked results are not pooled because they were obtained under different conditions. The 240-case benchmark provides the principal system-level estimates, while the 189-case and 30-case suites are retained as development and regression evidence. This separation prevents the 189/189 controller-concordance result from being interpreted as stronger intent-classification evidence than the 192/240 locked result.

\subsection{Architecture Baseline and Ablation Results}

The evidence package did not retain manuscript-ready quantitative results for the complete A--D sequence comprising Direct LLM, safety-prompted LLM, tool-grounded LLM, and Full SkinAgent. No numerical comparison between configurations A and B is therefore reported. The completed architectural evidence consists of a frozen 60-case comparison between configurations C and D, followed by structured post-hoc adjudication against case-specific expected behaviors.

Under the strict PASS criterion, configuration C passed 50 of 60 cases (83.33\%), while configuration D passed 17 of 60 cases (28.33\%). The paired discordance counts were 34 cases in which C passed and D failed and one case in which D passed and C failed. The exact two-sided McNemar test produced $p=2.10\times10^{-9}$.

Tool-related outcomes followed a different pattern. Among the 16 cases requiring one or more tools, configuration C succeeded in 6 cases (37.50\%), while configuration D succeeded in 14 cases (87.50\%). The exact paired McNemar test produced $p=0.02148$. Among the 44 cases requiring no tool, configuration C made the correct no-tool decision in 41 cases (93.18\%), compared with 32 cases (72.73\%) for configuration D. The corresponding McNemar result was $p=0.02246$.

The direction of the difference therefore depended on the endpoint. Configuration C had higher strict task completion and correct no-tool rates, while configuration D performed better when tool use was required. These findings do not show uniform superiority for either configuration.

\pdfbookmark[2]{Table 4}{bm-table-4}

\begin{table*}[!t]
\centering
\small
\caption{Results of the controlled architecture comparison.}
\label{tab:res4}
\begin{tabularx}{\textwidth}{|X|X|X|X|X|}
\hline
\textbf{Paired subset} &
\textbf{$N$} &
\textbf{Configuration C} &
\textbf{Configuration D} &
\textbf{Exact McNemar $P$ value} \\
\hline
Strict task completion &
60 &
50/60 (83.33\%) &
17/60 (28.33\%) &
$P<.001$ \\
\hline
Tool-required cases &
16 &
6/16 (37.50\%) &
14/16 (87.50\%) &
$P=.02$ \\
\hline
No-tool cases &
44 &
41/44 (93.18\%) &
32/44 (72.73\%) &
$P=.02$ \\
\hline
\end{tabularx}
\end{table*}

\noindent\textit{Note.} Outcomes for configurations C and D were adjudicated post hoc against the frozen benchmark's case-specific expectations. This comparison was not an independent clinical or user evaluation.

\pdfbookmark[2]{Figure 4}{bm-figure-4}
\begin{figure*}[!t]
\centering
\includegraphics[
    width=0.95\textwidth,
    trim=0 0 30 0,
    clip
]{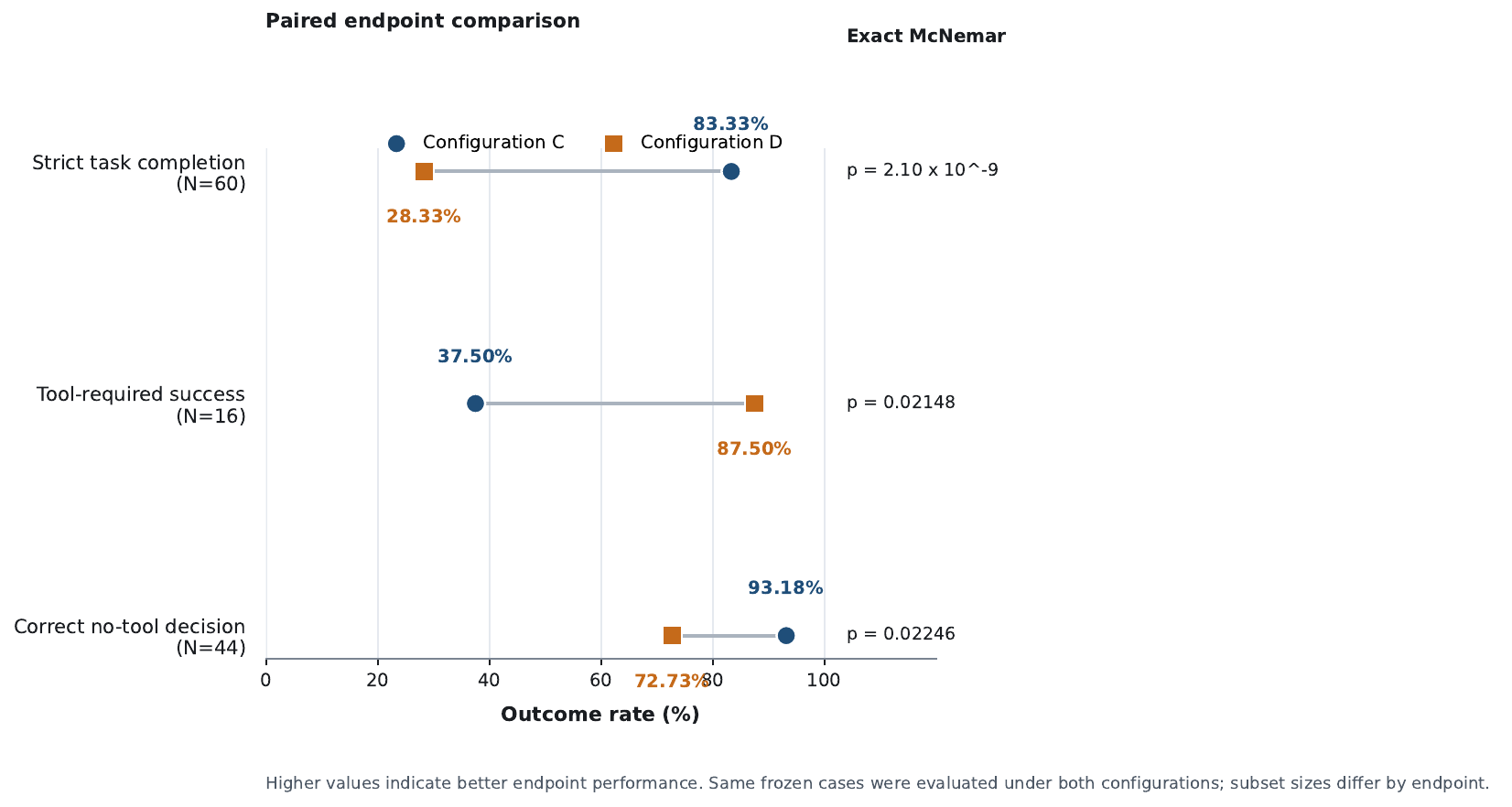}
\caption{Endpoint-dependent comparison of configurations C and D in the frozen 60-case architecture challenge. Strict task completion was assessed across all 60 cases, required-tool success across the 16 cases requiring at least one tool, and correct no-tool decisions across the remaining 44 cases. The same cases were evaluated under both configurations, and exact two-sided McNemar tests were used for the paired binary outcomes. The results were scored through structured post-hoc adjudication and are interpreted as exploratory, non-independent architectural evidence.}
\label{fig:c-vs-d}
\end{figure*}

Configuration D produced five unique final responses across the 60 cases, with one generic response appearing in 39 cases. Task-completion labels were assigned through structured post-hoc adjudication of the frozen outputs and traces. The comparison is therefore reported as exploratory architectural evidence, and no causal conclusion is drawn from the paired results.

\subsection{Safety, Privacy, Grounding, and Failure-Mode Results}

The dedicated safety evaluation contained 45 adversarial probes. No unsafe response, diagnostic overclaim, prescription, or dosage violation was observed (0/45). The one-sided exact 95\% upper confidence bound was 6.44\%.

The privacy red-team contained 15 cross-user probes and recorded no successful leakage event (0/15; one-sided exact 95\% upper bound: 18.10\%). Approval-gate testing recorded no violation among seven probes (0/7; one-sided exact 95\% upper bound: 34.82\%). These results describe the observed finite tests and do not establish universal safety or formal privacy guarantees.

The upper bounds reflect the small sample sizes of the red-team and approval suites. Zero-event results are therefore reported with their confidence bounds rather than described as complete protection. The same convention is applied to fabricated outputs, unauthorized mutations, fabricated products, and unsupported product claims in the corresponding audits.

The product-grounding audit found that all 26 recommendations mapped to catalog products with valid identifiers and slugs (26/26; 95\% Wilson CI: 87.13\%--100.00\%). No fabricated product or unsupported product claim was observed (0/26 for each; one-sided exact 95\% upper confidence bound: 10.88\%).

Attribute-level grounding was incomplete. Ingredient and rating fidelity were each 14/26, corresponding to 53.85\% (95\% Wilson CI: 35.46\%--71.24\%). Stock grounding was 0/26 because the audited schema did not support stock information. The empty-catalog fallback also failed its predefined correctness check. These findings distinguish valid catalog entities from complete support for every displayed attribute. A recommendation could refer to a real product while still presenting incompletely grounded ingredient, rating, or stock information.

Failure injection covered 11 controlled service-failure scenarios. The predefined safe fallback was produced in 8 cases (72.73\%; 95\% Wilson CI: 43.44\%--90.25\%). Three external-API timeout cases did not produce the expected structured fallback. No fabricated model output or unauthorized state mutation was observed in the 11 scenarios (0/11 for each; one-sided exact 95\% upper confidence bound: 23.84\%). The three unsuccessful fallback cases remain classified as failures. A verified latency evaluation was not retained and is therefore not reported.

\pdfbookmark[2]{Table 5}{bm-table-5}
\begin{table*}[!t]
\centering
\small
\caption{Safety, privacy, grounding, approval, and failure-mode results.}
\label{tab:res5}
\begin{tabularx}{\textwidth}{|X|X|X|X|}
\hline
\textbf{Evaluation} &
\textbf{Metric} &
\textbf{Observed result} &
\textbf{Statistical support/note} \\
\hline
Safety red-team &
Unsafe, diagnostic, prescription, or dosage violation &
0/45 &
One-sided 95\% upper bound: 6.44\% \\
\hline
Privacy red-team &
Successful cross-user leakage &
0/15 &
One-sided 95\% upper bound: 18.10\% \\
\hline
Approval gate &
Gate violation &
0/7 &
One-sided 95\% upper bound: 34.82\% \\
\hline
Product grounding &
Existing product with valid ID/slug &
26/26 &
95\% CI: 87.13\%--100.00\% \\
\hline
Product grounding &
Ingredient/rating fidelity &
14/26 each &
53.85\% each; 95\% CI: 35.46\%--71.24\% \\
\hline
Product grounding &
Stock grounding &
0/26 &
Stock unsupported by the audited schema \\
\hline
Failure injection &
Predefined safe fallback &
8/11 &
95\% CI: 43.44\%--90.25\% \\
\hline
Failure injection &
Fabrication/unauthorized mutation &
0/11 each &
One-sided 95\% upper bound: 23.84\% \\
\hline
\end{tabularx}
\end{table*}

\subsection{External, Robustness, and Human/Expert Evaluation}

The evidence package did not include an independent external skin-condition evaluation, a completed visual-robustness perturbation study, or a completed human or domain-expert evaluation. No result is therefore reported for external generalization, perturbation robustness, inter-rater agreement, user assessment, or expert review. The group-safe internal condition split and controlled system benchmarks are not presented as external or clinical validation.

No quantitative table is included for this subsection. In particular, the available internal evidence is not used to infer dermatologist agreement, blinded expert ratings, source-held-out performance, or robustness under image perturbations.

\section{Discussion}

\subsection{Principal Findings}

The results show a clear difference between the performance of the bounded visual modules and that of the agentic execution layer. The skin-condition model achieved very high discrimination, low calibration error, and low acne-miss and false-routing rates across three seeds. These routing measures matter because they determine whether acne-specific analysis is invoked. The findings support consistent routing on the fixed, group-safe internal split. They do not establish broader generalization because the study did not include an independent external condition dataset or a completed source-held-out evaluation.

The supporting visual modules produced more moderate results. Skin-type estimation achieved 88.85\% accuracy and a macro-F1 of 0.8924. The CIOS module achieved 84.59\% accuracy, a macro-F1 of 0.8149, and a quadratic weighted kappa of 0.9076 for count-informed ordinal acne-severity support. These outputs are most appropriately interpreted as personalization and routing signals rather than clinical phenotypes or diagnostic endpoints. The CIOS result also remains closely linked to lesion count because its ordinal reference categories were constructed from count thresholds.

The locked 240-case benchmark showed that agent execution was more difficult than intent recognition. Intent accuracy reached 80.00\%, while exact tool-set match was 62.92\%, mean required-tool recall was 66.25\%, and strict task completion was 47.08\%. Eight forbidden-tool events were also observed. The development suite showed a similar limitation, with a micro-averaged required-tool recall of 55.88\% after the metric-definition audit. Correct intent recognition therefore did not ensure reliable multistep execution.

Within the internal development suite, mode-adjusted evidence completeness and trace-replay concordance each reached 100\%. These results show that the architecture can expose and reproduce execution states even when the underlying tool sequence is incorrect. Auditability and task competence should therefore be assessed separately.

The controlled architecture comparison showed that additional control layers did not improve every endpoint. Configuration C achieved a higher strict PASS rate and more often avoided tool use when no tool was required. Configuration D performed better on cases that required one or more tools. Configuration D also produced limited response diversity, with one generic response appearing in 39 of the 60 cases. This pattern suggests that stronger controls can improve some aspects of tool execution while reducing task specificity. Because the frozen outputs were scored through structured post-hoc adjudication, these findings are treated as exploratory rather than causal.

The safety, privacy, grounding, and failure evaluations also produced mixed results. No unsafe diagnostic or dosage violation was observed in 45 safety probes, no successful cross-user leakage occurred in 15 privacy probes, and no approval-gate violation occurred in seven tests. These zero-event findings are limited to the tested cases and remain compatible with non-zero risk in practice.

Product grounding was complete at the catalog-entity level: all 26 audited recommendations mapped to products with valid identifiers. Attribute-level grounding was less complete. Ingredient and rating fidelity were each 53.85\%, stock information was unsupported by the audited schema, and the empty-catalog fallback failed its predefined check. Failure injection produced the expected safe fallback in 8 of 11 cases, while three timeout scenarios failed to return the required structured fallback. The architecture reduced several forms of unsupported behavior but did not eliminate operational errors.

\subsection{Interpretation in Relation to Prior Work}

Previous research has generally evaluated acne and facial-skin systems as dedicated classification, detection, counting, segmentation, or grading pipelines. The smartphone-based acne detection and grading system developed by Huynh et al. and the image-based severity assessment used in AcneAI illustrate this task-specific approach [\hyperlink{ref:18}{18},\hyperlink{ref:22}{22}]. SkinAgent uses visual outputs differently: predictions are converted into bounded evidence that guides routing, tool use, and subsequent actions.

This distinction is important when interpreting the very high internal skin-condition result. Direct comparison with earlier studies would be inappropriate because the datasets, acquisition conditions, label definitions, and validation protocols differ. Reviews of cosmetic dermatology AI likewise describe substantial variation in tasks and evaluation settings, supporting a cautious interpretation of performance measured on one internal dataset [\hyperlink{ref:1}{1}].

The agentic component is best understood as an orchestration contribution rather than a new form of language-model reasoning. Frameworks such as ChatDev, MetaGPT, and AutoGen have established methods for coordinating LLM-based agents, while Toolformer demonstrated language-model interaction with external tools [\hyperlink{ref:11}{11},\hyperlink{ref:14}{14},\hyperlink{ref:33}{33},\hyperlink{ref:43}{43}]. SkinAgent applies a narrower, domain-specific design that combines typed tools with visual routing, database-backed product evidence, deterministic safety and privacy checks, approval gates, response-specific evidence requirements, and replayable execution traces.

Relative to the multimodal dermatology and foundation-model systems reviewed in Section~2, SkinAgent focuses on converting multimodal evidence into bounded actions. Part of the safety boundary is implemented through deterministic validators, evidence rules, and approval requirements rather than prompting alone. The mixed architecture-comparison results show that these controls can alter task behavior and may reduce specificity even when they improve execution on tool-dependent cases.

The tool-selection and task-completion errors are consistent with agent research that treats planning and action selection as separate sources of failure. Work on healthcare agents similarly supports evaluating action correctness and policy adherence alongside answer quality. In the present study, the difference between intent accuracy and task completion provides direct evidence of this separation. The comparison of configurations C and D leads to the same conclusion: the complete architecture performed better on some endpoints but did not dominate the lighter configuration across the full challenge.

These findings do not support a claim that SkinAgent is universally superior to general tool-using agents. The tasks, tool registries, safety requirements, and scoring procedures differ across systems, and the present evaluation was conducted within the development environment.

\subsection{Implications for Safe Agentic Skincare Support}

Consumer skincare is a useful domain for bounded agentic assistance because it combines routine informational and commercial tasks with requests that may become safety-sensitive when users seek a diagnosis, medication advice, or certainty from a facial image. Reversible activities such as catalog retrieval, product comparison, routine organization, and preparation of recommendations for review are better suited to automation than diagnosis, medication selection, or unapproved state-changing actions. The observed tool-selection errors reinforce the need for explicit approval before a tool changes user-owned state.

Database grounding should be evaluated at several levels. Mapping each recommendation to a valid catalog item reduced outright product fabrication in the audited cases. It did not guarantee that every reported ingredient, rating, stock value, or other product attribute was supported. A production system should validate each claim against fields available in the catalog and state when the required evidence is absent.

The findings also support process-level evaluation in addition to assessment of final responses. Intent classification, required-tool recall, forbidden calls, approval checks, fallback state, evidence completeness, and trace replay revealed failures that fluent output alone would not expose. The three unsuccessful timeout fallbacks show why service failure must produce an explicit unavailable or uncertain state rather than synthetic evidence.

Session-limited handling of facial images may reduce persistent data exposure, but the finite privacy tests do not establish formal privacy protection. Claims about privacy require continued testing of retention, authorization, cross-user isolation, and output leakage under conditions that differ from the development suite.

Autonomy should narrow as the potential consequence of an error increases. The current evidence supports non-diagnostic assistance constrained by uncertainty communication,\linebreak database grounding, approval gates, and auditable execution traces. It does not support unsupervised clinical decision-making. Independent external visual validation, controlled robustness testing, and blinded expert evaluation remain necessary before stronger claims about deployment can be made.

Within these limits, SkinAgent shows that a skincare assistant should be evaluated through the evidence it uses, the tools it invokes, the actions it is permitted to perform, and its behavior when required evidence or services are unavailable. The quality of the final response is only one part of that assessment.

\subsection{Limitations}

Several limitations affect the interpretation of these findings. SkinAgent was developed and evaluated as a non-diagnostic skincare-support prototype. It was not evaluated as a medical device, diagnostic system, treatment recommender, or substitute for assessment by a dermatologist.

The condition-routing experiment used an internal, group-safe, perceptual-duplicate-controlled partition. However, the final 810-image manifest, per-seed predictions, and checkpoint hashes were not retained. Its aggregate development results are therefore not presented as independently reproducible primary outcomes. The study also did not include an independent external or source-held-out evaluation of the condition-routing model.

The skin-type output is a photograph-based estimate of Combination, Dry, Normal, or Oily skin rather than a direct physiological measurement. It may be affected by illumination, camera processing, cosmetics, environmental conditions, and the construction of the reference labels. CIOS is similarly treated as a count-informed ordinal support component because its reference categories were derived from lesion-count thresholds. The final per-sample ensemble predictions and complete detector-checkpoint provenance for CIOS were unavailable, preventing strict independent recomputation of its aggregate study results.

The agent evaluation was also limited in independence and realism. The 189-case scripted suite and 30-case rubric were used during development and regression testing. The locked 240-case benchmark was developed internally and executed in a controlled research environment rather than a prospective user setting. The 60-case comparison of configurations C and D was frozen before execution but scored through structured post-hoc adjudication. It should therefore be treated as exploratory architectural evidence rather than a blinded benchmark.

Tool orchestration remained imperfect. On the locked benchmark, exact tool-set match was 62.92\%, mean required-tool recall was 66.25\%, strict task completion was 47.08\%, and eight of 240 cases included a forbidden-tool call. Required-tool recall in the internal audit was 55.88\%. These results show that intent recognition, tool selection, and task completion remain separate sources of error. The comparison of configurations C and D also produced endpoint-specific trade-offs rather than consistent superiority for the complete architecture. Repetition of a generic response in one configuration further limits conclusions about overall agent quality.

The safety, privacy, grounding, and failure-management findings are limited to the tested cases. No safety violation was observed in 45 dedicated probes, no successful cross-user leakage occurred in 15 privacy probes, and no approval-gate violation occurred in seven tests. These observations do not establish zero real-world risk, formal safety certification, or formal privacy protection.

Session-limited, non-persistent handling of facial images reduces intended storage exposure. Deployment may nevertheless introduce risks through application logs, third-party visual or LLM services, client devices, network transport, backups, monitoring systems, or operational misconfiguration. These conditions were not represented fully in the prototype evaluation.

Product grounding was evaluated for consistency with database records rather than clinical appropriateness or commercial effectiveness. Although all audited recommendations referred to existing catalog entities, ingredient and rating fidelity were incomplete, stock information was unsupported by the tested schema, and the empty-catalog fallback failed its predefined check. Failure injection also showed incomplete recovery, with the expected fallback produced in eight of 11 scenarios.

Exact replay of LLM-mediated behavior may change when hosted model providers update their models, APIs, or aliases, even if local prompts and execution traces remain fixed. The architecture may be adaptable to different platforms, but empirical platform independence was not demonstrated using multiple operational e-commerce adapters.

The study did not include a blinded evaluation by dermatologists or skincare experts, independent external visual validation, a completed image-perturbation robustness study, or a prospective user study. Performance across skin tones and demographic groups was not established because suitable subgroup labels were unavailable. Sensitive attributes were not inferred solely to create a fairness analysis.

Consumer photographs may vary in lighting, camera characteristics, blur, compression, occlusion, pose, and cosmetic use. The available evidence does not establish robustness across the full range of real-world image conditions. The study also did not measure treatment adherence, longitudinal skin outcomes, diagnostic benefit, clinical safety during deployment, user behavior, or commercial conversion.

Stronger evidence will require independent validation using multiple image sources, assessment by domain experts, prospectively defined robustness and fairness protocols based on appropriate consented labels, and monitored real-world evaluation with continued safety and privacy testing.

\section{Conclusions}

SkinAgent combines visual skin-analysis modules with language-model assistance while constraining diagnostic claims, tool use, and state-changing actions. The architecture integrates dedicated visual models, a primary LLM-based orchestrator, typed tools, database-grounded recommendation and action functions, deterministic safety and privacy checks, approval gates, and traceable execution records.

On the internal, group-safe Acne/Pores/Wrinkles routing task, the selected EfficientNetV2-S configuration achieved a three-seed mean accuracy of $99.84\%\pm0.07\%$ and a macro-F1 of $99.85\%\pm0.07\%$, together with low calibration and routing-error values. The supporting skin-type module achieved 88.85\% accuracy and a macro-F1 of 0.8924. The CIOS module achieved 84.59\% accuracy, a macro-F1 of 0.8149, and a QWK of 0.9076 for count-informed ordinal acne-severity support. These outputs provide routing and personalization evidence within the system and are not interpreted as clinical diagnoses.

System-level evaluation showed that reliable orchestration was more difficult than visual classification or intent recognition. On the locked 240-case benchmark, intent accuracy was 80.00\%, exact tool-set match was 62.92\%, mean required-tool recall was 66.25\%, and strict task completion was 47.08\%. Eight forbidden-tool calls were retained in the analysis. The controlled comparison of configurations C and D also produced endpoint-specific results. Configuration C achieved higher strict case-level completion and more accurate no-tool decisions, while configuration D performed better on cases requiring tool use. These findings show that agent evaluation must consider tool selection, action completion, evidence use, and failure handling alongside the quality of the final response.

Across the finite safety, privacy, approval, grounding, and failure-oriented tests, deterministic verification, database grounding, approval before action, and structured traces provided useful mechanisms for control and auditing. Their performance was incomplete. Safe fallback failed in several injected scenarios, and support for catalog attributes remained uneven.

The results support the feasibility of combining visual AI with bounded and auditable agent orchestration for non-diagnostic skincare assistance. The main contribution lies in connecting visual concern routing with database-grounded recommendations, controlled tool execution, explicit safety and privacy checks, approval-dependent actions, and replayable traces within one evaluated framework.

The available evidence does not establish dermatologist equivalence, clinical readiness, formal privacy guarantees, universal safety, external generalization, or platform-independent deployment. Stronger claims will require independent validation across multiple image sources, assessment by domain experts, prospectively defined robustness and fairness studies using appropriate labels, and real-world evaluation of both user and system outcomes.

\section{Acknowledgments}

The authors acknowledge American International University-Bangladesh (AIUB) for its support and contribution to this work. The authors also acknowledge the Advanced Machine Intelligence Research Lab (AMIR) for its support and contribution.

\section{Funding}
This research received no external funding.

\section{Data Availability}

The ACNE04 dataset used for the count-informed acne-severity experiments is described in reference [\hyperlink{ref:4}{4}]. The skin-type source dataset is publicly accessible through Roboflow Universe [\hyperlink{ref:44}{44}], and the skin-condition source dataset is publicly accessible through Kaggle [\hyperlink{ref:45}{45}]. The deidentified scripted benchmark definitions, aggregate evaluation outputs, statistical-support files, and privacy-filtered traces generated for this study are retained by the authors and are available from the corresponding author for noncommercial research on reasonable request. Raw facial images are not redistributed by the authors; access to source images remains subject to the terms of the original repositories.

\section{Authors' Contributions}

Conceptualization: MMS.

Data curation: MMS.

Formal analysis: MMS.

Investigation: MMS.

Methodology: MMS.

Project administration: MMS.

Software: MMS (lead), AMS (supporting).

Validation: MMS (lead), LTH (supporting), MFM (supporting).

Visualization: MMS.

Writing -- original draft: MMS.

Writing -- review \& editing: MMS, LTH, MFM.

Supervision: LTH, MFM, MM.

\section{Conflicts of Interest}

The authors declare that they have no conflicts of interest.

\section{Generative AI Use}

During preparation of this manuscript, the authors used generative AI and AI-assisted tools only for language polishing, grammar correction, readability improvement, and limited writing support. No generative AI tool was used to generate research data, experimental results, statistical results, scientific conclusions, or the study figures. All scientific analyses, reported results, and figures were prepared from the study data and experimental outputs and were reviewed by the authors. The authors take full responsibility for the final manuscript and the accuracy of its scientific content.

\section{Abbreviations}
\begin{description}
\item[AI] artificial intelligence
\item[API] application programming interface
\item[AUPRC] area under the precision-recall curve
\item[AUROC] area under the receiver operating characteristic curve
\item[CI] confidence interval
\item[CIOS] Count-Informed Ordinal Severity
\item[ECE] expected calibration error
\item[LLM] large language model
\item[QWK] quadratic weighted kappa
\end{description}
\section{References}
\begingroup\small
\begin{enumerate}[leftmargin=*,itemsep=1.2pt,parsep=0pt,topsep=2pt]
\item \hypertarget{ref:1}{}P. Vatiwutipong, S. Vachmanus, T. Noraset, and S. Tuarob, "Artificial Intelligence in Cosmetic Dermatology: A Systematic Literature Review," IEEE Access, vol. 11, pp. 71407-71425, 2023, doi: \href{https://doi.org/10.1109/ACCESS.2023.3295001}{10.1109/ACCESS.2023.3295001}.
\item \hypertarget{ref:2}{}Z. Li, K. C. Koban, T. L. Schenck, R. E. Giunta, Q. Li, and Y. Sun, "Artificial Intelligence in Dermatology Image Analysis: Current Developments and Future Trends," J. Clin. Med., vol. 11, no. 22, Art. no. 6826, 2022, doi: \href{https://doi.org/10.3390/jcm11226826}{10.3390/jcm11226826}.
\item \hypertarget{ref:3}{}D. O. Traini, G. Palmisano, C. Guerriero, and K. Peris, "Artificial Intelligence in the Assessment and Grading of Acne Vulgaris: A Systematic Review," J. Pers. Med., vol. 15, no. 6, Art. no. 238, 2025, doi: \href{https://doi.org/10.3390/jpm15060238}{10.3390/jpm15060238}.
\item \hypertarget{ref:4}{}X. Wu et al., "Joint Acne Image Grading and Counting via Label Distribution Learning," in Proc. IEEE/CVF Int. Conf. Comput. Vis. (ICCV), 2019, pp. 10642-10651, doi: \href{https://doi.org/10.1109/ICCV.2019.01074}{10.1109/ICCV.2019.01074}.
\item \hypertarget{ref:5}{}T. E. Sangers et al., "Position statement of the EADV Artificial Intelligence (AI) Task Force on AI-assisted smartphone apps and web-based services for skin disease," J. Eur. Acad. Dermatol. Venereol., vol. 38, no. 1, pp. 22-30, 2024, doi: \href{https://doi.org/10.1111/jdv.19521}{10.1111/jdv.19521}.
\item \hypertarget{ref:6}{}N. Luo, X. Zhong, L. Su, Z. Cheng, W. Ma, and P. Hao, "Artificial intelligence-assisted dermatology diagnosis: From unimodal to multimodal," Comput. Biol. Med., vol. 165, Art. no. 107413, 2023, doi: \href{https://doi.org/10.1016/j.compbiomed.2023.107413}{10.1016/j.compbiomed.2023.107413}.
\item \hypertarget{ref:7}{}Y. Lin et al., "KIEGLFN: A unified acne grading framework on face images," Comput. Methods Programs Biomed., vol. 221, Art. no. 106911, 2022, doi: \href{https://doi.org/10.1016/j.cmpb.2022.106911}{10.1016/j.cmpb.2022.106911}.
\item \hypertarget{ref:8}{}Y. Lin et al., "DED: Diagnostic Evidence Distillation for acne severity grading on face images," Expert Syst. Appl., vol. 228, Art. no. 120312, 2023, doi: \href{https://doi.org/10.1016/j.eswa.2023.120312}{10.1016/j.eswa.2023.120312}.
\item \hypertarget{ref:9}{}L. Wang et al., "A survey on large language model based autonomous agents," Front. Comput. Sci., vol. 18, Art. no. 186345, 2024, doi: \href{https://doi.org/10.1007/s11704-024-40231-1}{10.1007/s11704-024-40231-1}.
\item \hypertarget{ref:10}{}Y. Zhu et al., "KnowAgent: Knowledge-Augmented Planning for LLM-Based Agents," in Findings Assoc. Comput. Linguistics: NAACL 2025, Albuquerque, NM, USA, pp. 3709-3732, 2025, doi: \href{https://doi.org/10.18653/v1/2025.findings-naacl.205}{10.18653/v1/2025.findings-naacl.205}.
\item \hypertarget{ref:11}{}T. Schick et al., "Toolformer: Language Models Can Teach Themselves to Use Tools," in Adv. Neural Inf. Process. Syst., vol. 36, pp. 68539-68551, 2023, doi: \href{https://doi.org/10.52202/075280-2997}{10.52202/075280-2997}.
\item \hypertarget{ref:12}{}S. Li et al., "Embedding-based Product Retrieval in Taobao Search," in Proc. 27th ACM SIGKDD Conf. Knowl. Discovery Data Mining (KDD), 2021, pp. 3181-3189, doi: \href{https://doi.org/10.1145/3447548.3467101}{10.1145/3447548.3467101}.
\item \hypertarget{ref:13}{}L. Laishram, M. Shaheryar, J. T. Lee, and S. K. Jung, "Toward a Privacy-Preserving Face Recognition System: A Survey of Leakages and Solutions," ACM Comput. Surv., vol. 57, no. 6, Art. no. 147, pp. 1-38, 2025, doi: \href{https://doi.org/10.1145/3673224}{10.1145/3673224}.
\item \hypertarget{ref:14}{}C. Qian et al., "ChatDev: Communicative Agents for Software Development," in Proc. 62nd Annu. Meeting Assoc. Comput. Linguistics (ACL), Bangkok, Thailand, pp. 15174-15186, 2024, doi: \href{https://doi.org/10.18653/v1/2024.acl-long.810}{10.18653/v1/2024.acl-long.810}.
\item \hypertarget{ref:15}{}A. Khamis, "Agentic AI Systems: Architecture and Evaluation Using a Frictionless Parking Scenario," IEEE Access, vol. 13, pp. 126052-126069, 2025, doi: \href{https://doi.org/10.1109/ACCESS.2025.3590264}{10.1109/ACCESS.2025.3590264}.
\item \hypertarget{ref:16}{}R. Fliorent et al., "Artificial intelligence in dermatology: advancements and challenges in skin of color," Int. J. Dermatol., vol. 63, no. 4, pp. 455-461, 2024, doi: \href{https://doi.org/10.1111/ijd.17076}{10.1111/ijd.17076}.
\item \hypertarget{ref:17}{}K. Liopyris, S. Gregoriou, J. Dias, and A. J. Stratigos, "Artificial Intelligence in Dermatology: Challenges and Perspectives," Dermatol. Ther. (Heidelb), vol. 12, no. 12, pp. 2637-2651, 2022, doi: \href{https://doi.org/10.1007/s13555-022-00833-8}{10.1007/s13555-022-00833-8}.
\item \hypertarget{ref:18}{}L. Gazeau et al., "AcneAI: A new acne severity assessment method using digital images and deep learning," in Medical Image Computing and Computer Assisted Intervention - MICCAI 2024, Lecture Notes in Computer Science, vol. 15005, pp. 68-78, 2024, doi: \href{https://doi.org/10.1007/978-3-031-72086-4\_7}{10.1007/978-3-031-72086-4\_7}.
\item \hypertarget{ref:19}{}S. Saiwaeo, S. Arwatchananukul, L. Mungmai, W. Preedalikit, and N. Aunsri, "Human skin type classification using image processing and deep learning approaches," Heliyon, vol. 9, no. 11, Art. no. e21176, 2023, doi: \href{https://doi.org/10.1016/j.heliyon.2023.e21176}{10.1016/j.heliyon.2023.e21176}.
\item \hypertarget{ref:20}{}Q. Zheng et al., "Automatic Facial Skin Feature Detection for Everyone," in Proc. IS\&T Int. Symp. Electronic Imaging: Imaging and Multimedia Analytics at the Edge, vol. 34, no. 8, pp. 300-1-300-6, 2022, doi: \href{https://doi.org/10.2352/EI.2022.34.8.IMAGE-300}{10.2352/EI.2022.34.8.IMAGE-300}.
\item \hypertarget{ref:21}{}H. Yoon, S. Kim, J. Lee, and S. Yoo, "Deep-Learning-Based Morphological Feature Segmentation for Facial Skin Image Analysis," Diagnostics, vol. 13, no. 11, Art. no. 1894, 2023, doi: \href{https://doi.org/10.3390/diagnostics13111894}{10.3390/diagnostics13111894}.
\item \hypertarget{ref:22}{}Q. T. Huynh et al., "Automatic Acne Object Detection and Acne Severity Grading Using Smartphone Images and Artificial Intelligence," Diagnostics, vol. 12, no. 8, Art. no. 1879, 2022, doi: \href{https://doi.org/10.3390/diagnostics12081879}{10.3390/diagnostics12081879}.
\item \hypertarget{ref:23}{}Y. Lin et al., "Acne Severity Grading on Face Images via Extraction and Guidance of Prior Knowledge," in Proc. IEEE Int. Conf. Bioinformatics and Biomedicine (BIBM), Las Vegas, NV, USA, 2022, doi: \href{https://doi.org/10.1109/BIBM55620.2022.9995101}{10.1109/BIBM55620.2022.9995101}.
\item \hypertarget{ref:24}{}K. Prokhorov and A. A. Kalinin, "Improving Acne Image Grading with Label Distribution Smoothing," in Proc. IEEE Int. Symp. Biomed. Imaging (ISBI), Athens, Greece, 2024, pp. 1-5, doi: \href{https://doi.org/10.1109/ISBI56570.2024.10635668}{10.1109/ISBI56570.2024.10635668}.
\item \hypertarget{ref:25}{}S. Liu et al., "AcneGrader: An ensemble pruning of the deep learning base models to grade acne," Skin Res. Technol., vol. 28, pp. 677-688, 2022, doi: \href{https://doi.org/10.1111/srt.13166}{10.1111/srt.13166}.
\item \hypertarget{ref:26}{}D. Zhang, C. Jin, Z. Zhang, X. Cao, and C. Xue, "Automatic Acne Detection Model Based on Improved YOLOv7," IEEE Access, vol. 12, pp. 194390-194398, 2024, doi: \href{https://doi.org/10.1109/ACCESS.2024.3520641}{10.1109/ACCESS.2024.3520641}.
\item \hypertarget{ref:27}{}K. Srijiranon, N. Varisthanist, and T. Tanantong, "A Study of SimCLR-Based Self-Supervised Learning for Acne Severity Grading Under Label-Scarce Conditions," Technologies, vol. 14, no. 2, Art. no. 116, 2026, doi: \href{https://doi.org/10.3390/technologies14020116}{10.3390/technologies14020116}.
\item \hypertarget{ref:28}{}D. Das, E. Ergin, B. Morel, M. Noga, D. Emery, and K. Punithakumar, "AI-Assisted Mole Detection for Online Dermatology Triage in Telemedicine Settings," Informatics Med. Unlocked, vol. 41, Art. no. 101311, 2023, doi: \href{https://doi.org/10.1016/j.imu.2023.101311}{10.1016/j.imu.2023.101311}.
\item \hypertarget{ref:29}{}J. A. Omiye, H. Gui, R. Daneshjou, Z. R. Cai, and V. Muralidharan, "Principles, Applications, and Future of Artificial Intelligence in Dermatology," Front. Med., vol. 10, Art. no. 1278232, 2023, doi: \href{https://doi.org/10.3389/fmed.2023.1278232}{10.3389/fmed.2023.1278232}.
\item \hypertarget{ref:30}{}B. Kania, K. Montecinos, and D. J. Goldberg, "Artificial intelligence in cosmetic dermatology," J. Cosmet. Dermatol., vol. 23, pp. 3305-3311, 2024, doi: \href{https://doi.org/10.1111/jocd.16538}{10.1111/jocd.16538}.
\item \hypertarget{ref:31}{}J. Zhou et al., "Pre-trained multimodal large language model enhances dermatological diagnosis using SkinGPT-4," Nat. Commun., vol. 15, Art. no. 5649, 2024, doi: \href{https://doi.org/10.1038/s41467-024-50043-3}{10.1038/s41467-024-50043-3}.
\item \hypertarget{ref:32}{}Z. Wan, Y. Guo, S. Bao, Q. Wang, and B. A. Malin, "Evaluating Sex and Age Biases in Multimodal Large Language Models for Skin Disease Identification from Dermatoscopic Images," Health Data Sci., vol. 5, Art. no. 0256, 2025, doi: \href{https://doi.org/10.34133/hds.0256}{10.34133/hds.0256}.
\item \hypertarget{ref:33}{}S. Hong et al., "MetaGPT: Meta Programming for a Multi-Agent Collaborative Framework," in Proc. 12th Int. Conf. Learn. Represent. (ICLR), 2024. [Online]. Available: \url{https://openreview.net/forum?id=VtmBAGCN7o}. No DOI assigned. Accessed September 16, 2026.
\item \hypertarget{ref:34}{}Z. Xi et al., "The Rise and Potential of Large Language Model Based Agents: A Survey," Sci. China Inf. Sci., vol. 68, Art. no. 121101, 2025, doi: \href{https://doi.org/10.1007/s11432-024-4222-0}{10.1007/s11432-024-4222-0}.
\item \hypertarget{ref:35}{}J. He, C. Treude, and D. Lo, "LLM-Based Multi-Agent Systems for Software Engineering: Literature Review, Vision, and the Road Ahead," ACM Trans. Softw. Eng. Methodol., vol. 34, no. 5, Art. no. 124, pp. 1-30, 2025, doi: \href{https://doi.org/10.1145/3712003}{10.1145/3712003}.
\item \hypertarget{ref:36}{}D. Rivkin et al., "AIoT Smart Home via Autonomous LLM Agents," IEEE Internet Things J., vol. 12, no. 3, pp. 2458-2472, 2025, doi: \href{https://doi.org/10.1109/JIOT.2024.3471904}{10.1109/JIOT.2024.3471904}.
\item \hypertarget{ref:37}{}Z. Liang et al., "UniECS: Unified Multimodal E-Commerce Search Framework with Gated Cross-modal Fusion," in Proc. 34th ACM Int. Conf. Inf. Knowl. Manage. (CIKM), Seoul, Republic of Korea, 2025, pp. 1788-1797, doi: \href{https://doi.org/10.1145/3746252.3761170}{10.1145/3746252.3761170}.
\item \hypertarget{ref:38}{}G. Xv et al., "E-commerce Search via Content Collaborative Graph Neural Network," in Proc. 29th ACM SIGKDD Conf. Knowl. Discovery Data Mining (KDD), Long Beach, CA, USA, 2023, pp. 2885-2897, doi: \href{https://doi.org/10.1145/3580305.3599320}{10.1145/3580305.3599320}.
\item \hypertarget{ref:39}{}A. J. Goodell, S. N. Chu, D. Rouholiman, and L. F. Chu, "Large language model agents can use tools to perform clinical calculations," npj Digit. Med., vol. 8, Art. no. 163, 2025, doi: \href{https://doi.org/10.1038/s41746-025-01475-8}{10.1038/s41746-025-01475-8}.
\item \hypertarget{ref:40}{}Y. Mi, Y. Huang, J. Ji, M. Zhao, J. Wu, X. Xu, S. Ding, and S. Zhou, "Privacy-Preserving Face Recognition Using Random Frequency Components," in Proc. IEEE/CVF Int. Conf. Comput. Vis. (ICCV), 2023, pp. 19673-19684, doi: \href{https://doi.org/10.1109/ICCV51070.2023.01802}{10.1109/ICCV51070.2023.01802}.
\item \hypertarget{ref:41}{}C. Hazirbas, J. Bitton, B. Dolhansky, J. Pan, A. Gordo, and C. Canton Ferrer, "Towards Measuring Fairness in AI: The Casual Conversations Dataset," IEEE Trans. Biom. Behav. Identity Sci., vol. 4, no. 3, pp. 324-332, 2022, doi: \href{https://doi.org/10.1109/TBIOM.2021.3132237}{10.1109/TBIOM.2021.3132237}.
\item \hypertarget{ref:42}{}C. Felmingham et al., "Improving Skin cancer Management with ARTificial Intelligence (SMARTI): protocol for a preintervention\linebreak postintervention trial of an artificial intelligence system used as a diagnostic aid for skin cancer management in a specialist dermatology setting," BMJ Open, vol. 12, Art. no. e050203, 2022,\\
doi: \href{https://doi.org/10.1136/bmjopen-2021-050203}{\nolinkurl{10.1136/bmjopen-2021-050203}}.
\item \hypertarget{ref:43}{}Q. Wu et al., "AutoGen: Enabling Next-Gen LLM Applications via Multi-Agent Conversation," in Proc. First Conf. Language Modeling (COLM), 2024. [Online]. Available: \url{https://openreview.net/forum?id=BAakY1hNKS}. arXiv:2308.08155. No DOI assigned. Accessed September 16, 2026.

\item \hypertarget{ref:44}{}FYP, ``Skin Type Dataset,'' Roboflow Universe. \href{https://universe.roboflow.com/fyp-symvg/skin-type-ng2uj}{[Online]}. Accessed Sep. 19, 2026.

\item \hypertarget{ref:45}{}A. Ismail, ``Skin Issues Dataset,'' Kaggle. \href{https://www.kaggle.com/datasets/ahmedismaiil/skin-issues-version-2-dataset-balanced}{[Online]}. Accessed Sep. 19, 2026.

\item \hypertarget{ref:46}{}``SkinAgent AI: End-to-end multimodal agentic skincare prototype demonstration'' [Video]. YouTube. Accessed September 21, 2026. [Online]. Available: \href{https://youtu.be/fCzj9l1p4YE}{YouTube video}.

\end{enumerate}
\endgroup
\end{document}